\documentclass[final,5p, times,twocolumn, a4paper]{elsarticle}
\usepackage{amssymb}
\usepackage{lipsum}
\usepackage{multirow}
\usepackage{adjustbox}
\usepackage{array}
\usepackage{subfigure}
\usepackage{balance}
\usepackage{amsmath,amssymb,amsfonts}
\usepackage{graphicx}
\usepackage{textcomp}
\usepackage{comment}
\usepackage{subcaption}
\usepackage{caption}
\usepackage{float}
\usepackage{booktabs}
\usepackage{tcolorbox} 
\usepackage{parskip}
\usepackage{booktabs}
\usepackage{colortbl}
\usepackage{url}
\usepackage{soul}
\usepackage{pifont}
\usepackage{ragged2e}
\usepackage{algpseudocode}
\usepackage[utf8]{inputenc}
\usepackage{amsmath}
\usepackage{hyperref}
\usepackage[ruled,vlined]{algorithm2e}
\SetKwComment{Comment}{$\triangleright$\ }{}

\makeatletter
\newcommand{\removelatexerror}{\let\@latex@error\@gobble}
\makeatother

\journal{arxiv}

\begin{document}

\begin{frontmatter}

\title{An Efficient Deep Learning Framework for Brain Stroke Diagnosis Using Computed Tomography Images}

\author[bu]{Md. Sabbir Hossen} \ead{sabbir.hossen@bu.edu.bd}
\author[bu]{Eshat Ahmed Shuvo} \ead{mdshuvoslk012@gmail.com}
\author[msu]{Shibbir Ahmed Arif} \ead{arifs1@montclair.edu}
\author[bu,mbstu]{Pabon Shaha} \ead{pabonshahacse15@gmail.com}
\author[gsu,niter]{Anichur Rahman } \ead{ar36248@georgiasouthern.edu}
\author[bu]{Md. Saiduzzaman} 
\ead{sojib.cse56bu@gmail.com}
\author[bsbi]{Fahmid Al Farid  } \ead{fahmid.alfarid@berlinsbi.com }
\author[mu]{Hezerul Abdul Karim } \ead{hezerul@mmu.edu.my }
\author[aizu]{Abu Saleh Musa Miah } \ead{musa@u-aizu.ac.jp }

\address[bu]{Department of Computer Science and Engineering, Bangladesh University, Mohammadpur, \\ Dhaka - 1207, Bangladesh}
\address[msu]{School of Computing, Montclair State University, Montclair, NJ 07043, USA}
\address[mbstu]{Department of Computer Science and Engineering, Mawlana Bhashani Science and Technology University,\\ Tangail-1902, Dhaka, Bangladesh}
\address[gsu]{School of Computing, Georgia Southern University, Statesboro, GA 30458, Georgia, USA }
\address[niter]{Department of Computer Science and Engineering, National Institute of Textile Engineering and Research (NITER), \\Constituent Institute of Dhaka University, Savar, Dhaka-1350, Bangladesh}
\address[bsbi]{Faculty of Computer Science and Informatics, Berlin School of Business and Innovation  Karl-Marx-Straße 97-99,Berlin,  12043, Germany}
\address[mu]{Centre for Image and Vision Computing (CIVC), COE for Artificial Intelligence, Faculty of Artificial Intelligence and Engineering (FAIE), Multimedia University, Cyberjaya 63100, Malaysia }
\address[aizu]{Department of Computer Science and Engineering, The University of Aizu, Aizuwakamatsu, Fukushima, Japan }

\begin{abstract}
Brain stroke is a leading cause of mortality and long-term
disability worldwide, underscoring the need for precise and rapid prediction techniques. A Computed Tomography (CT) scan is considered one of the most effective methods for diagnosing brain strokes. Most stroke classification techniques use a single slice-level
prediction mechanism, requiring radiologists to manually select the most critical CT slice from the original CT volume. Although clinical evaluations are often used in traditional diagnostic procedures, machine learning (ML) has opened up new avenues for improving stroke diagnosis. To supplement traditional diagnostic techniques, this study investigates machine learning models for early brain stroke prediction using CT scan images. This research proposes a novel machine learning approach to brain stroke detection, focusing on optimizing classification performance with pre-trained deep learning models and advanced optimization strategies. Pre-trained models, including DenseNet201, InceptionV3, MobileNetV2, ResNet50, and Xception, are used for feature extraction. Feature engineering techniques, including BFO, PCA, and LDA, further enhance model performance. These features are then classified using machine learning algorithms, including SVC, RF, XGB, DT, LR, KNN, and GNB. Our experiments demonstrate that the combination of MobileNetV2, LDA, and SVC achieved the highest classification accuracy of 97.93\%, significantly outperforming other model-optimizer-classifier combinations. The results underline the effectiveness of integrating lightweight pre-trained models with robust optimization and classification techniques for brain stroke diagnosis. \vspace{3mm}
\end{abstract}


\end{frontmatter}

\section{Introduction}
\label{introduction}

Brain Stroke is a medical condition, one of the leading causes of disability and death worldwide, and is brought on by a disruption of blood supply to the brain \cite{murphy}. Brain stroke occurs when a blockage or ruptured blood vessel interrupts blood flow to the brain, causing oxygen deprivation that damages brain cells. Most of the brain strokes are either ischemic or hemorrhagic \cite{grysiewicz}. An ischemic stroke occurs when a blood clot or blockage prevents blood flow to the brain, depriving brain tissue of oxygen and nutrients and ultimately resulting in the death of brain cells \cite{feske2021ischemic}. A hemorrhagic stroke occurs when a weak blood vessel ruptures, causing an internal or external brain hemorrhage. It is frequently associated with high blood pressure or aneurysms. This kind of stroke directly damages the brain due to the pressure of accumulated blood \cite{montano2021hemorrhagic}.  

Stroke is the second most common cause that contributes significantly to the global mortality burden \cite{STROKEAHA}. Stroke resulted in over 143 million disability-adjusted life years (DALYs) and 6.6 million deaths in 2019 \cite{feigin2021global}. The age-standardized prevalence rate (ASPR) for stroke was 1,240.263 per 100,000 people. Hemorrhagic strokes had an ASPR of 350.338 per 100,000 people, whereas ischemic strokes, which accounted for around 80\% of all stroke occurrences, had an ASPR of 950.973 per 100,000 people \cite{zhang2024global}. 

Stroke prevention and prompt response are vital public health priorities because of the tremendous burden this puts on families and healthcare systems throughout the globe. Prevention of brain stroke involves managing risk factors such as high blood pressure, diabetes, obesity, and smoking through lifestyle changes, regular medical check-ups, and medication \cite{caplan}. Unfortunately, many of the symptoms or risk factors for stroke can be mild or develop gradually, making it difficult to identify those who are at risk, especially in the early stages \cite{elendu}. The chance of a stroke can be considerably decreased by early identification and suitable therapies, however traditional approaches are reactive rather than predictive and frequently depend on patient-reported symptoms. 

Brain stroke can be identified using various conventional methods, which mostly rely on clinical information and patient histories, magnetic resonance imaging (MRI), and computed tomography (CT) scans \cite{ABBASI2023100145}. In recent years, Machine learning (ML) has emerged as an effective instrument in the diagnosis, treatment, and prognosis of diseases, including stroke. Machine learning can identify brain stroke early using both image data and categorical data \cite{daidone}. However, many researchers in the past utilized Machine Learning to detect brain stroke by using CT scans, MRI, ECG, or categorical data; for instance, \cite{patel} \cite{polamuri} used CT images and deep learning techniques for classifying brain stroke. \cite{abulfaraj} \cite{al2022deep} developed deep learning models to improve early brain stroke identification using MRI images. \cite{mondal} \cite{bathla} made use of the Stroke Prediction categorical dataset to detect brain stroke. \cite{choi2021deep} used raw ECG sensor data to detect stroke at an early stage. 

According to the study above, researchers used different kinds of datasets, such as MRI, CT scan, ECG, or Categorical data, to diagnose brain stroke. However, Computed Tomography (CT) Scans and Magnetic Resonance Imaging (MRI) are the most widely used and reliable methods \cite{kwong2003computed}. Deep learning and machine learning hold promise for technological advancement in challenging medical scenarios, especially in brain stroke diagnosis. There are some improvements needed in brain stroke classification using machine learning with CT scan data. This research presents a method that uses a CT scan image dataset to detect brain stroke early and prevent mortality and disability by combining many machine-learning approaches. Our major contributions are as follows.

\begin{itemize}
    \item This research created a more robust dataset for brain stroke diagnosis by merging and curating existing datasets. This dataset addresses potential gaps in publicly available CT scan datasets. 
    \item We utilized several feature extraction methods, including ResNet50, DenseNet201, InceptionV3, Xception, and MobileNetV2.
    \item Three feature engineering techniques, PCA, LDA, and BFO, were employed.
    \item Seven traditional classifiers were used: SVC, DT, RF, GNB, XGB, KNN, and LR.
    \item Accuracy, Precision, Recall, and F1 score were utilized to evaluate the performance of the proposed model.
\end{itemize}

The remaining sections of this paper are organized as follows: Section II provides a thorough literature review and examines the drawbacks of existing approaches. Section III reviews the dataset and techniques used for machine learning-based brain stroke detection and provides a detailed explanation of data preparation and feature extraction. Section IV describes the experimental setup and implementation. Section V critically evaluates the results of combining several machine learning techniques. Section VI concludes the paper, outlines potential future applications, and emphasizes the project's significance.

\section{Literature Review}
Numerous researchers have introduced brain stroke detection models using traditional and machine learning techniques. However, this field remains relatively underdeveloped, and ongoing research aims to address its limitations. To gain a clearer understanding of the current state of knowledge in this area, several existing studies that used computed tomography (CT) scans to diagnose brain stroke have been reviewed and presented here for further analysis. \textit{A. Abumihsan et al.} \cite{abumihsan} used a hybrid feature extraction method with a convolutional block attention module (CBAM) to introduce a novel method for ischemic stroke identification from CT scan images. The hybrid feature extraction method provides a thorough representation of brain CT images by leveraging the advantages of two pre-trained models, DenseNet121 and MobileNetV3, through feature fusion. A specialized, first-hand dataset (Dataset 1) collected from the Specialized Private Hospital in Palestine was used to design and assess the proposed methodology. The approach was further tested on a publicly available dataset (Dataset 2) to demonstrate its generalizability and robustness. The findings demonstrate that the suggested model performed exceptionally well, achieving an accuracy of 99.21\% on Dataset 1 and 98.73\% on Dataset 2. However, this study is limited to only two feature extraction methods, which may restrict feature diversity and model performance. \textit{M.M Hossain et al.} \cite{hossainmaruf} suggested a novel hybrid ViT-LSTM model to accurately detect stroke using CT scan images. Additionally, they applied Explainable AI (XAI) methods like SHAP, attention mapping, and LIME to provide trustworthy and reasonable predictions and assure clinical relevance. They evaluated the model using BrSCTHD-2023 dataset and the Kaggle brain stroke dataset. The proposed ViT-LSTM model performed the best with 94.55\% accuracy with layer normalization and 92.22\% accuracy without layer normalization using BrSCTHD-2023 dataset. Besides, it achieved 96.61\% accuracy on the Kaggle dataset. Despite this, the hybrid ViT-LSTM model requires higher computational cost, making real-time deployment in clinical settings more challenging. 
\begin{table*}[htbp]
\centering
\scriptsize
\setlength{\tabcolsep}{36pt}
\renewcommand{\arraystretch}{1}
\caption{Overview of State-of-the-Art Approaches and Their Limitations on Brain Stroke Diagnosis Using CT Images}
\label{tab:my-table}
\begin{tabular}{@{}llll@{}}
\toprule
\textbf{Ref.} &
  \textbf{Model} &
  \textbf{\begin{tabular}[c]{@{}c@{}}Accuracy\end{tabular}} &
  \textbf{Limitations} \\ \midrule
\cite{abumihsan} &
  Hybrid CBAM&
  \begin{tabular}[c]{@{}c@{}}99.21\%\end{tabular} &
  \begin{tabular}[c]{@{}c@{}}Limited to only two feature extraction methods.\end{tabular} \\ 
\cite{hossainmaruf} &
  \begin{tabular}[c]{@{}c@{}}ViT-LSTM \end{tabular} &
  \begin{tabular}[c]{@{}c@{}}96.61\%\end{tabular} &
  \begin{tabular}[c]{@{}c@{}}Higher computational cost making real-time deployment challenging.\end{tabular} \\
\cite{tahyudin2025high} &
  \begin{tabular}[c]{@{}c@{}}CBAM+ResNet18\end{tabular} &
  \begin{tabular}[c]{@{}c@{}}95.00\%\end{tabular} &
  \begin{tabular}[c]{@{}c@{}}Used small and imbalanced dataset, may contain bias and limit reliability.\end{tabular} \\ 
\cite{umamaheswaran2024enhanced} &
  \begin{tabular}[c]{@{}c@{}}XGB\end{tabular} &
  \begin{tabular}[c]{@{}c@{}}97.00\%\end{tabular} &
  \begin{tabular}[c]{@{}c@{}} Comparisons were made with older models.\end{tabular} \\ 
\cite{saleem} &
  BiLSTM&
  \begin{tabular}[c]{@{}c@{}}96.50\%\end{tabular} &
  \begin{tabular}[c]{@{}c@{}}Decision making process is lacking in explainability.\end{tabular} \\ 
\cite{sabir} &
  DCNN &
  96.50\% &
  \begin{tabular}[c]{@{}c@{}}Trained with limited data.\end{tabular} \\
\cite{prasher2024brain} &
  EfficientNet-B0 &
  98.72\% &
  \begin{tabular}[c]{@{}c@{}}Lack of diversity in the data samples.\end{tabular} \\ 
\cite{kulathilake} &
  \begin{tabular}[c]{@{}c@{}}DenseNet201\end{tabular} &
  \begin{tabular}[c]{@{}c@{}}98.02\%\end{tabular} &
  \begin{tabular}[c]{@{}c@{}}Experiment conducted in only one architecture\end{tabular} \\
\cite{raj2023strokevit} &
  StrokeViT &
  92.00\% &
  \begin{tabular}[c]{@{}c@{}}Clinical validation is limited.\end{tabular} \\ 
\cite{diker} &
  \begin{tabular}[c]{@{}c@{}}VGG19\end{tabular} &
  \begin{tabular}[c]{@{}c@{}}97.06\%\end{tabular} &
  \begin{tabular}[c]{@{}c@{}}Limited to a specific and small
dataset.\end{tabular} \\  \bottomrule
\end{tabular} \vspace{-2mm}
\end{table*}
\textit{I. Tahyudin et al.} \cite{tahyudin2025high} developed a deep learning model incorporating Convolutional Block Attention Module (CBAM) with ResNet18 to classify stroke in brain CT images. They used a publicly available dataset containing 2501 brain CT
images from patients, with 950 neurological stroke instances and 1551 healthy, non-stroke instances. The CBAM-ResNet18 model outperformed the standard ResNet18 model with 95\% validation accuracy. It also obtained an AUC score of 0.99, demonstrating its capability to classify the stroke and non-stroke samples well. Regardless of its outstanding performance, it has some drawbacks, such as generalization issues and limited data size. \textit{SK. UmaMaheswaran et al.} \cite{umamaheswaran2024enhanced} presented a novel hybrid technique to detect acute strokes from CT scan images. They used hybrid preprocessing approaches to enhance the quality of images. After that, they used discrete wavelet transform, Gabor Filter, and local binary pattern to extract the features from the segmented images. They then used the dingo optimization algorithm to select optimal features from the extracted images. Afterward, the selected optimal features were input into a novel ML model to detect normal or acute strokes. The proposed XGBoost model achieved an accuracy of 97\%. \textit{M.A. Saleem et al.} \cite{saleem} developed detection system using CT brain images combining a genetic algorithm to select the most relevant features and a bidirectional long short-term memory (BiLSTM) model to detect brain strokes at the early stage. The BiLSTM model achieved an accuracy of 96.5\%. They also used cross-validation to evaluate the model's overall performance using precision, recall, F1 score, ROC, and AUC. In addition to that, they compared the model’s performance with classical ML models such as SVM, LR, NB, DT, and RF, where the BiLSTM model outperformed all of them. However, the model's decision-making process is lacking in explainability. \textit{M. Sabir et al.} \cite{sabir} introduced a novel deep convolutional neural network (DCNN) model for early-stage brain stroke detection using CT scan images. The model consists of a feature extractor, a feature fusion module, and a stroke detection module. The proposed model performed better than state-of-the-art models like VGG16, ResNet50, and InceptionV3, achieving higher accuracy, sensitivity, and specificity. In their observation, the proposed model has obtained an accuracy of 96.5\% to detect stroke within 6 hours of onset. Despite this, the model was trained with limited data, which might have affected its generalizability. \textit{S. Prasher et al.} \cite{prasher2024brain} developed a deep learning model based on EfficientNet-B0 CNN architecture to predict brain strokes from 2515 CT scan images. The proposed model performed impressively, achieving an accuracy of 98.72\%. Thus, they demonstrated its ability to help doctors detect brain strokes from CT images. The researchers think that the model might better identify strokes by being trained with more CT images to provide better diagnoses to stroke patients. However, there is a lack of diversity in the data samples. \textit{C.D. Kulathilake et al.} \cite{kulathilake} presents a multiclass brain stroke detection model using the NCCT dataset, leveraging two deep learning models based on the DenseNet201 transfer learning architecture. The proposed approach follows a two-step process to identify ischemic stroke and its subtypes. To enhance convergence speed and improve training efficiency, the 1-cycle policy and FastAI framework were integrated into the models. Model 01 achieved a training accuracy of 97.03\%, a validation accuracy of 94.50\%. Similarly, Model 02 attained a training accuracy of 98.02\%, a validation accuracy of 95.67\%. The DenseNet201 architecture, combined with FastAI techniques, demonstrated superior performance compared to traditional and task-specific algorithms.  However, this research experiment used only one architecture and did not explore other architectures. \textit{R. Raj et al.} \cite{raj2023strokevit} proposed a hybrid framework, StrokeViT with AutoML, combining CNNs, Vision Transformers (ViT), and AutoML to improve stroke diagnosis from CT scan images. In this framework, the ViTs were used to extract long-range dependencies, whereas the CNNs were used to capture the local features to improve the slice-wise predictions. AutoML generated patient-level predictions by integrating slice-level data, achieving an accuracy of 87\% for the slice-level data as well as successfully classified the patient-level data with 92\% accuracy, which surpassed the standalone architectures by an accuracy of 9\%. Authors claimed that this technique can adapt for different diseases diagnosed in musculoskeletal and cancer conditions. Nonetheless, the clinical validation is limited in the proposed model. \textit{A. Diker et al.} \cite{diker} introduced a deep learning architecture for categorizing strokes using the Brain Stroke CT scan Dataset. For testing and training purposes, 2501 computed tomography (CT) images of brain strokes were employed in the experimental investigation. This study aimed to categorize brain stroke CT scans as normal and as stroke using a number of well-known pre-trained convolutional neural networks (CNNs), including GoogleNet, AlexNet, VGG-16, VGG-19, and Residual CNN. With an accuracy of 97.06\%, VGG-19 obtained the highest score. Though the results are promising, they are limited to a specific and small dataset. 
\begin{figure*}[htbp]
	\centering 
	\fbox{\includegraphics[height=10cm, width=18cm]{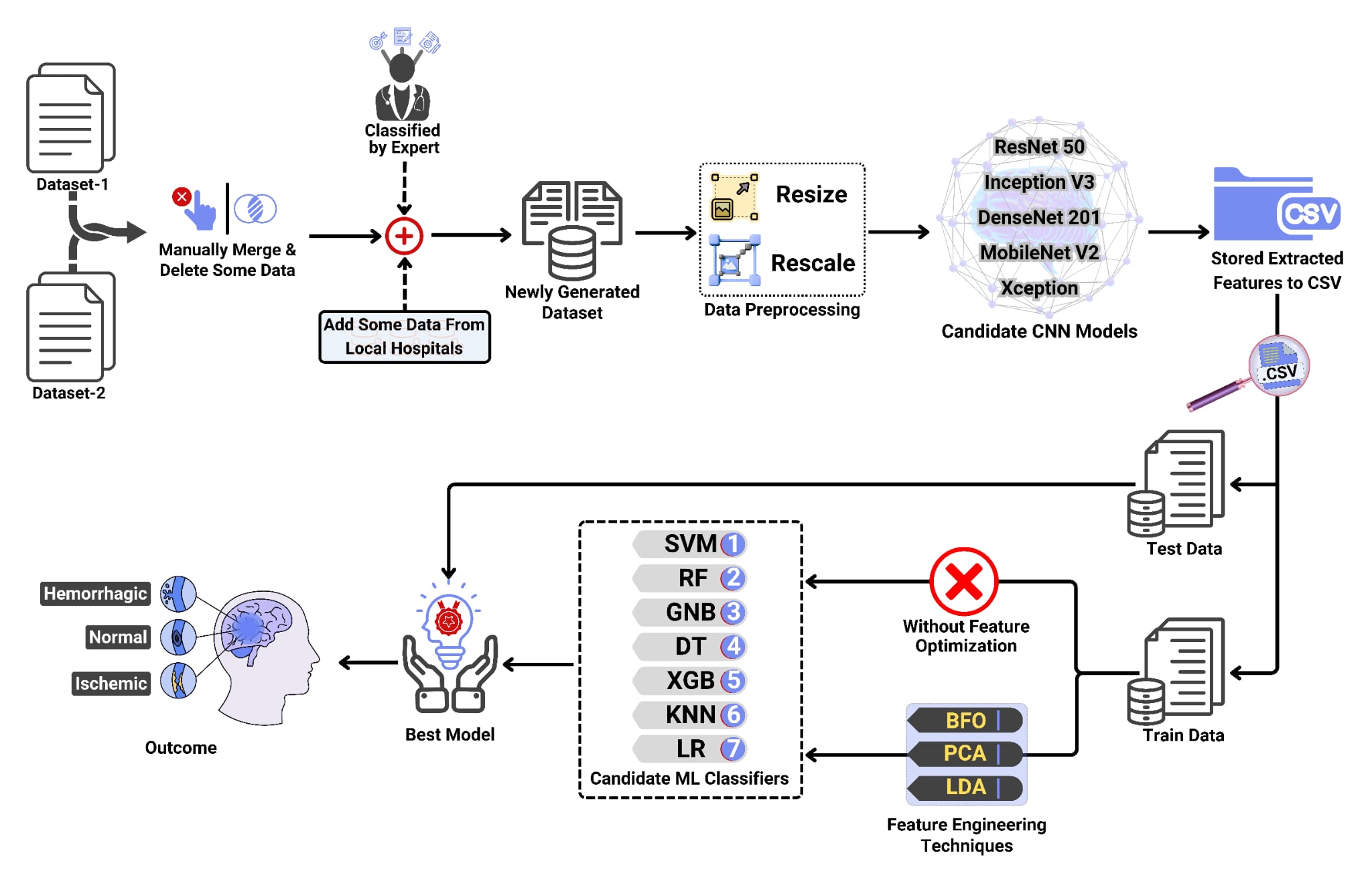}}	
		\caption{Proposed Architecture for Brain Stroke Detection} 
	\label{fig:md}
\end{figure*}
In conclusion, extensive research has been conducted on brain stroke detection using deep learning and CNN-based models. However, most existing studies rely on computationally expensive, heavyweight architectures and typically address only binary classification problems, differentiating stroke from non-stroke cases. This binary approach overlooks the clinical significance of distinguishing between different types of strokes, such as ischemic and hemorrhagic. To address these limitations, our study proposes a lightweight hybrid deep learning model designed to reduce computational complexity while maintaining high diagnostic accuracy. Additionally, we introduce a multi-class CT scan dataset comprising ischemic stroke, hemorrhagic stroke, and normal cases, which enables more clinically relevant and fine-grained classification. \vspace{-2mm}

\section{Materials \& Methods}
The methodology of this study is meticulously designed to develop an efficient and robust system for brain stroke detection and classification using CT scan images. By leveraging advanced deep learning techniques and integrating diverse datasets, the study aims to address the challenges of accurately identifying normal, ischemic, and hemorrhagic brain conditions. A systematic approach was adopted, involving data preprocessing and model architecture development, to ensure a comprehensive and reliable framework for stroke diagnosis. Fig. \ref{fig:md} illustrates the architectural diagram of this research, and this section provides a detailed explanation of the processes and techniques implemented throughout the study.

\subsection{Dataset Description}
The data utilized in this research to detect brain strokes were gathered from a public source and a local clinic. We initially collected two datasets. The name of Dataset 1 is the Brain Stroke CT Image Dataset \cite{dataset2501}, which contains 2501 images, and the name of Dataset 2 is the brain-stroke-prediction-ct-scan-image-dataset \cite{dataset2515}, which consists of 2515 images. Together, the two datasets consist of 5016 data. However, Dataset 1 comprises CT scan images of stroke and normal brain data, whereas Dataset 2 consists of CT scan images of ischemic and hemorrhagic strokes. Initially, both datasets contained binary classifications. With the help of an expert, we merged the two datasets to construct a three-class dataset comprising Normal, Ischemic, and Hemorrhagic categories. This integration involved incorporating all data from Dataset 2 while selectively including only the normal brain data from Dataset 1. During this process, we removed 424 instances from the original 1,551 normal brain images to enhance the dataset's quality and relevance. Additionally, we supplemented the dataset by incorporating 177 new normal instances obtained from a local clinic in Bangladesh. 
\begin{table}[htbp]
\setlength{\tabcolsep}{6pt}
\renewcommand{\arraystretch}{1.2}
\caption{Final Dataset of CT Scan Images for Brain Stroke Detection}
\centering
\begin{tabular}{ccccc}
\hline
      \textbf{Class }              & \textbf{Training} & \textbf{Testing} & \textbf{Validation} & \textbf{Total} \\ \hline
Normal     & 758            & 359           & 187                 & 1304  \\ 
Ischemic   & 928           & 401           & 222                 & 1551  \\ 
Hemorrhagic & 711            & 159           & 94                  & 964  \\ \hline
\end{tabular}
\label{tab:dt}
\end{table}
\begin{figure*}[]
\centering
\subfigure[Normal]{\label{fig:a}\includegraphics[width=29.3mm]{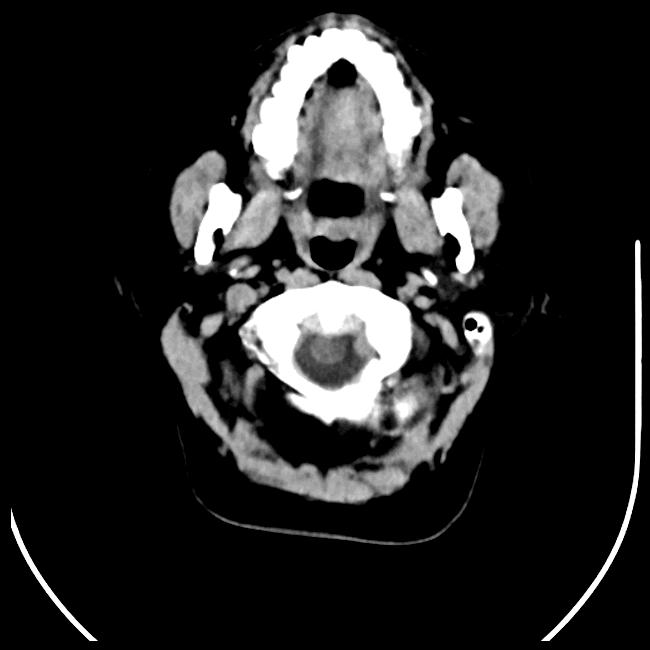}}
\subfigure[Normal]{\label{fig:b}\includegraphics[width=29.3mm]{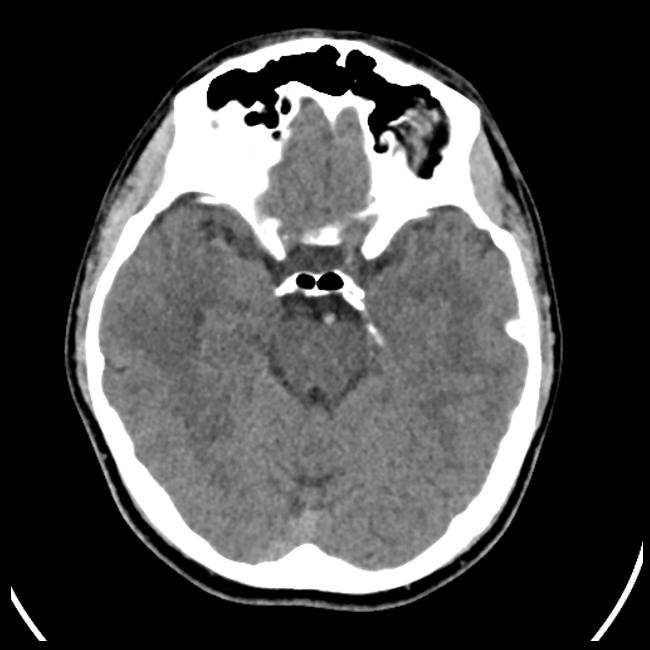}}
\subfigure[Normal]{\label{fig:c}\includegraphics[width=29.3mm]{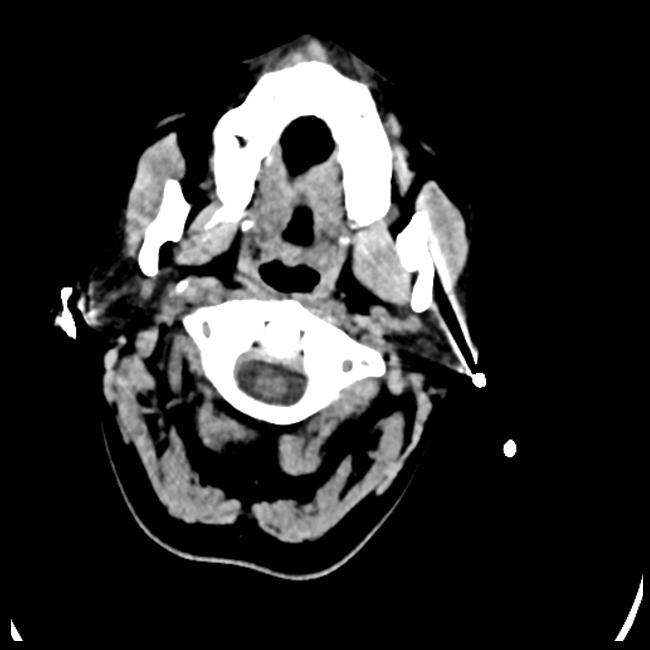}}
\subfigure[Normal]{\label{fig:d}\includegraphics[width=29.3mm]{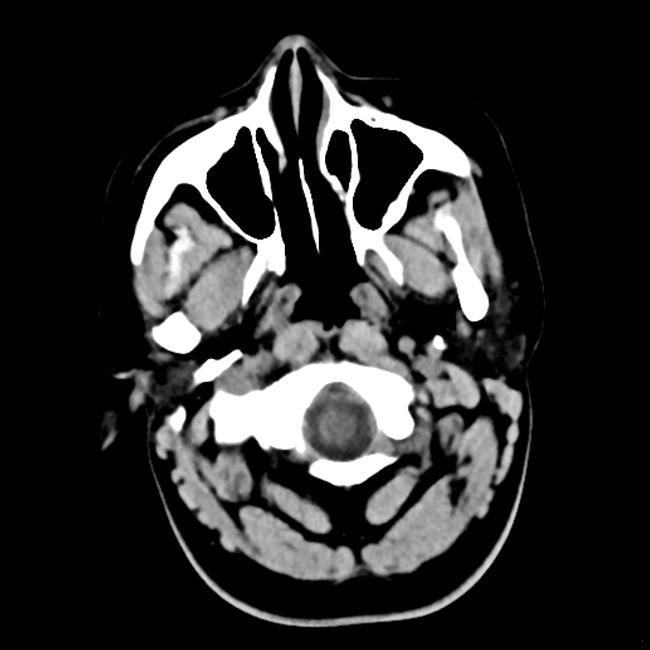}}
\subfigure[Normal]{\label{fig:e}\includegraphics[width=29.3mm]{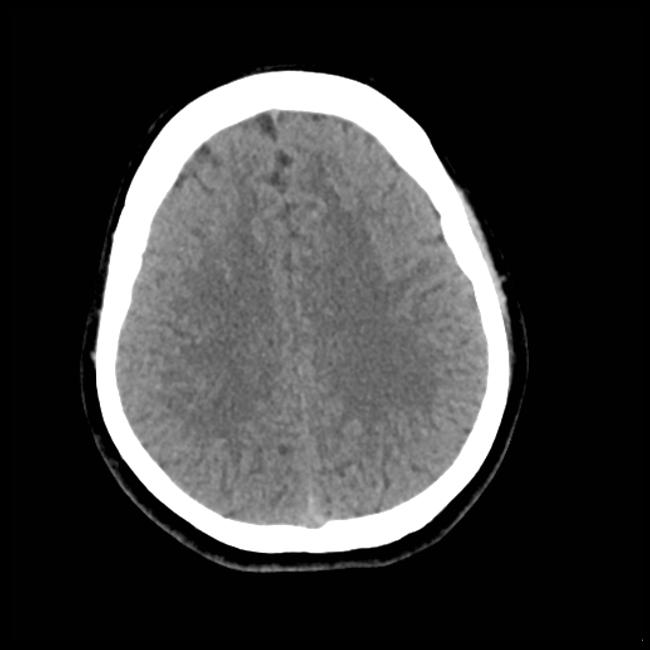}}
\subfigure[Normal]{\label{fig:f}\includegraphics[width=29.3mm]{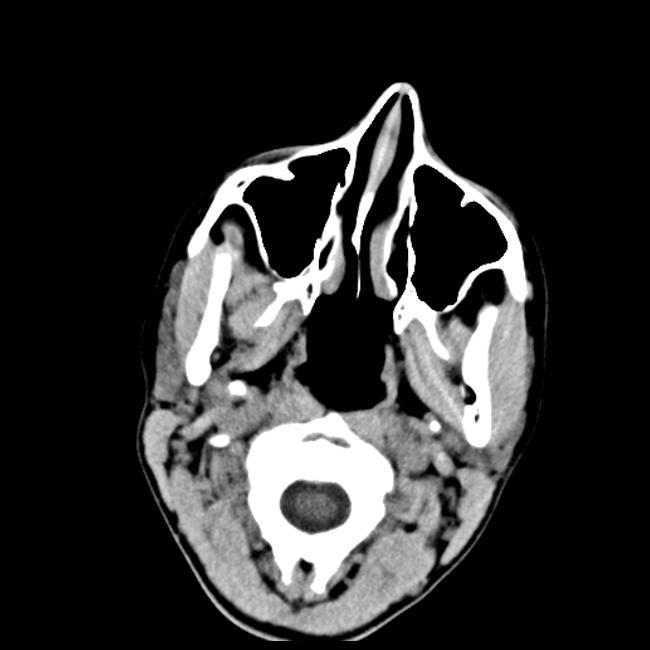}}
\subfigure[Hemorrhagic]{\label{fig:g}\includegraphics[width=29.3mm]{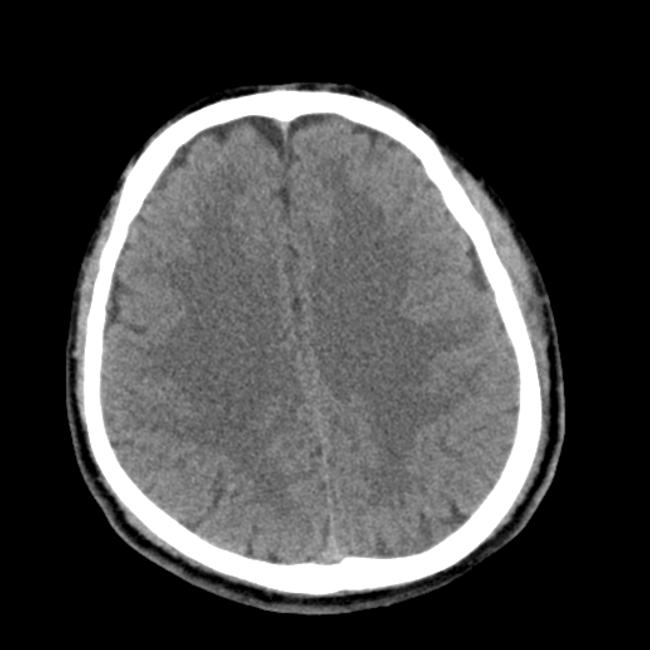}}
\subfigure[Hemorrhagic]{\label{fig:h}\includegraphics[width=29.3mm]{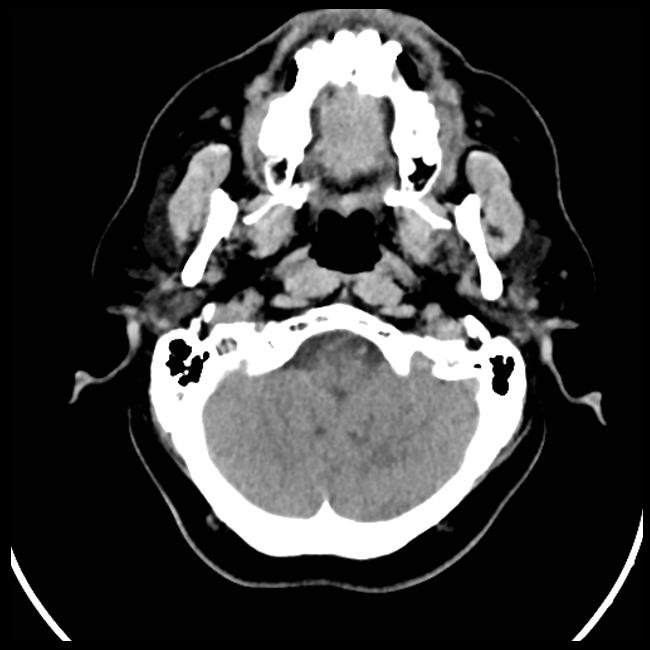}}
\subfigure[Hemorrhagic]{\label{fig:i}\includegraphics[width=29.3mm]{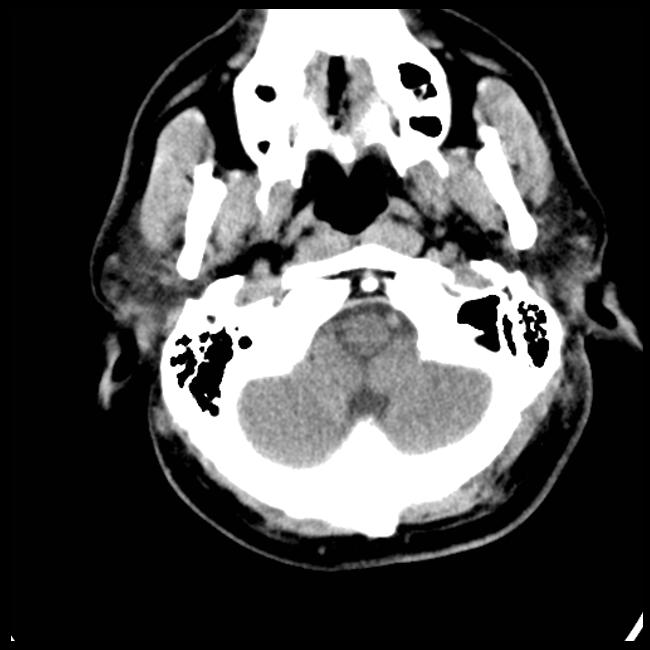}}
\subfigure[Hemorrhagic]{\label{fig:j}\includegraphics[width=29.3mm]{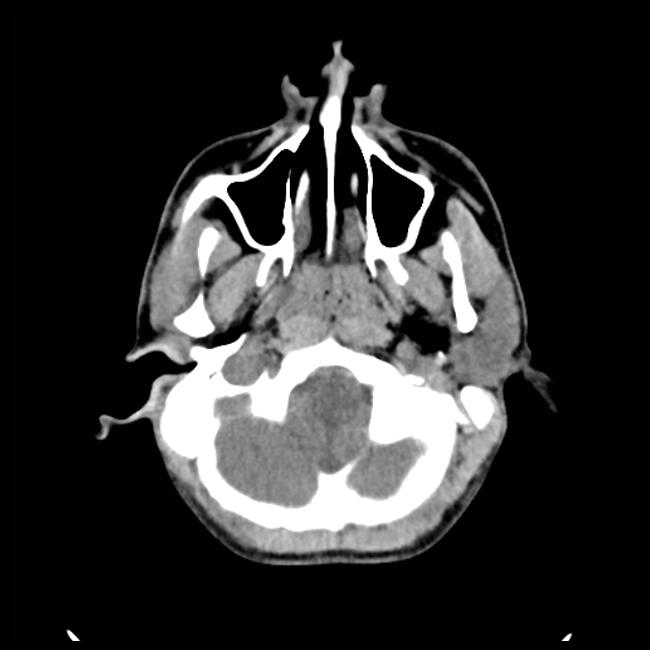}}
\subfigure[Hemorrhagic]{\label{fig:k}\includegraphics[width=29.3mm]{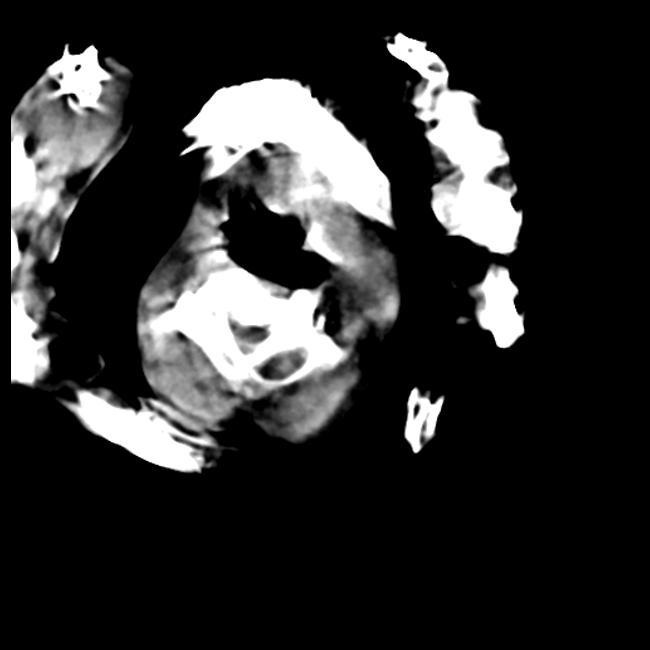}}
\subfigure[Hemorrhagic]{\label{fig:l}\includegraphics[width=29.3mm]{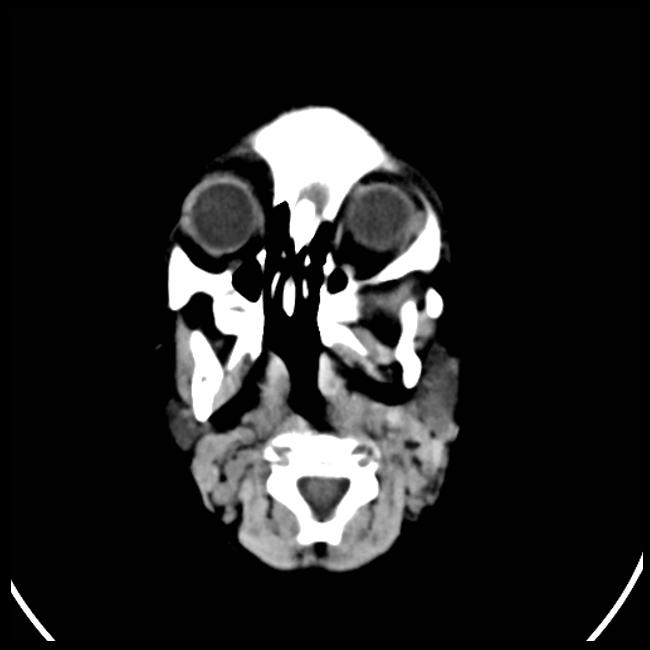}}
\subfigure[Ischemic]{\label{fig:m}\includegraphics[width=29.3mm]{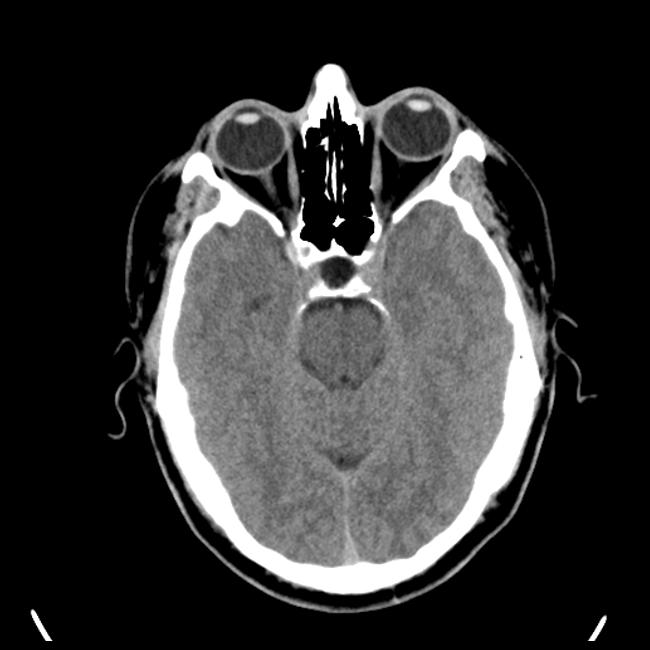}}
\subfigure[Ischemic]{\label{fig:n}\includegraphics[width=29.3mm]{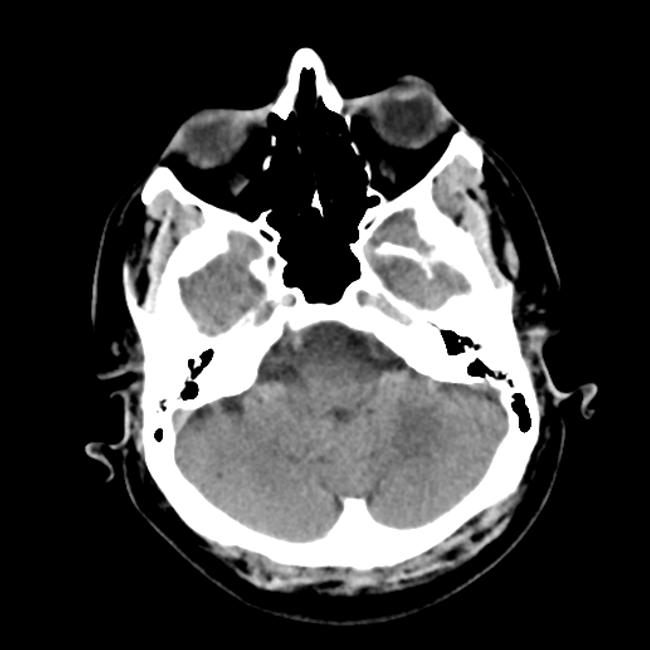}}
\subfigure[Ischemic]{\label{fig:o}\includegraphics[width=29.3mm]{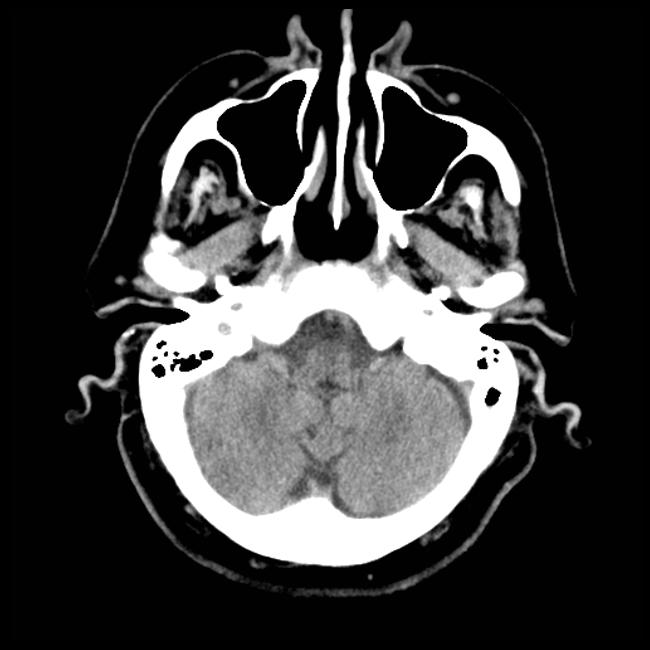}}
\subfigure[Ischemic]{\label{fig:p}\includegraphics[width=29.3mm]{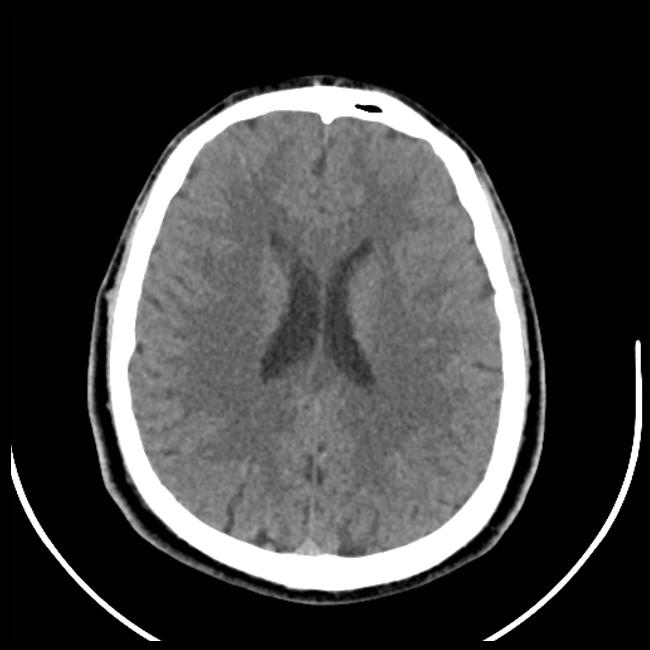}}
\subfigure[Ischemic]{\label{fig:q}\includegraphics[width=29.3mm]{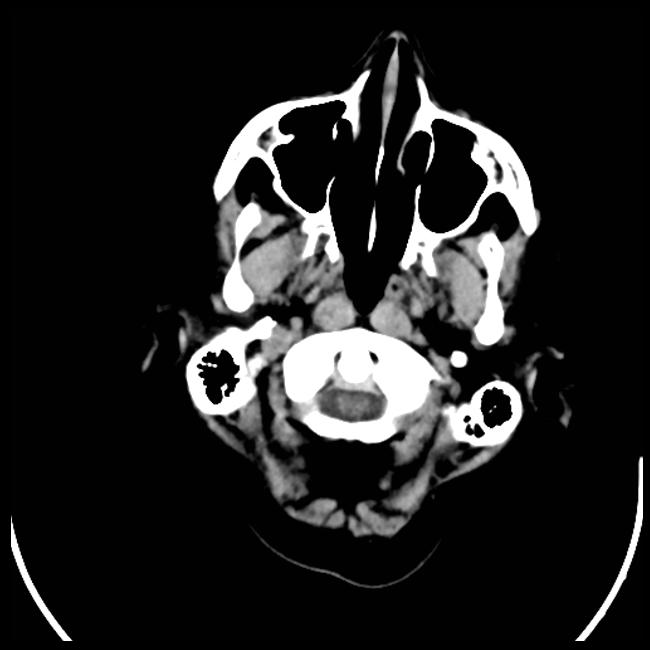}}
\subfigure[Ischemic]{\label{fig:r}\includegraphics[width=29.3mm]{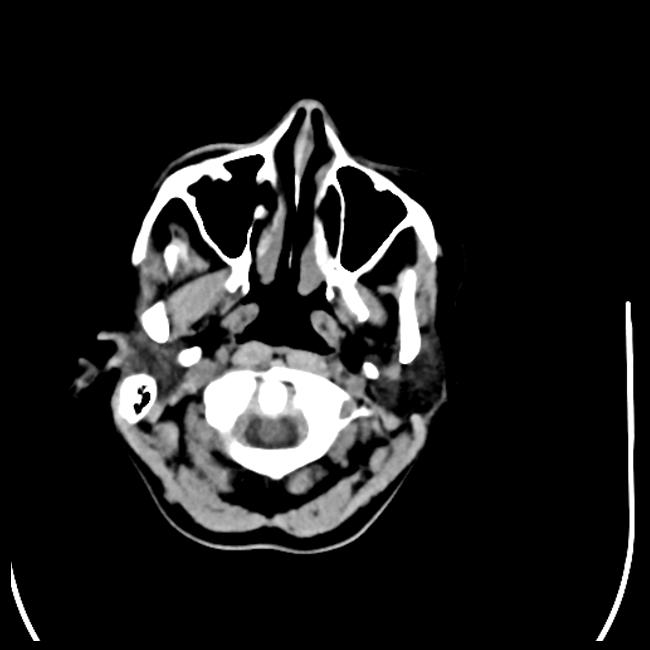}}
\caption{Sample images of Computed Tomography (CT) scan in normal, hemorrhagic, and Ischemic states. \vspace{-3mm}}
\label{fig:bi}
\end{figure*}
The newly generated dataset consists of 3819 CT scan images, split into three categories: Train, Test, and Validation. There are 2397 images in the training set, 919 in the test set, and 503 in the validation set. The three categories of normal, hemorrhagic, and ischemic images are applied to each of these groups, as shown in Table \ref{tab:dt}. Samples of each class of brain stroke images are displayed in Fig. \ref{fig:bi}  

\subsection{Image Preprocessing}
Building a machine learning model requires data preprocessing since noise may be present in the images. Preprocessing techniques are essential for eliminating noisy data from the dataset and achieving high training accuracy. Preprocessing is necessary to increase efficiency since training a convolutional neural network directly with raw images may result in poor classification performance. A number of preprocessing techniques were used on the images before they were added to the model. This study included resizing and rescaling as pre-processing techniques. Convolutional Neural Networks require input images that have consistent sizes. Consequently, each image in the dataset was resized to have consistent dimensions of 224x224 pixels. Also we used data augmentation method. The technique of artificially creating additional data from preexisting data is known as data augmentation \cite{strawberry}.  Through various adjustments like as rotation, flipping, scaling, and cropping, these techniques generate new images from the previously collected dataset. The model becomes less prone to overfitting and more resilient with the addition of new data, which finally improves the model's performance on unobserved data.

\subsection{Feature Extraction}
Feature extraction involves identifying essential patterns or characteristics from raw data to make it suitable for machine learning models \cite{featureextraction3}. In deep learning, particularly with convolutional neural networks (CNNs), this process applies convolutional filters to input data to detect features like edges, textures, shapes, and high-level object structures. Each layer in the network captures increasingly abstract features, starting from simple patterns (e.g., lines and edges) in the initial layers to complex features (e.g., object parts or relationships) in deeper layers \cite{featureextraction1}. The process employs convolutions, pooling for dimensionality reduction, and activation functions to create a compact and meaningful representation of the input, enabling efficient data processing for classification or detection tasks \cite{featureextraction2}. 

In this research, we utilize pre-trained deep learning models, DenseNet201, InceptionV3, MobileNetV2, ResNet50, and Xception, to extract discriminative features from CT scan images. The process begins with image preprocessing, where images are resized to match the input dimensions required by each model (e.g., 224×224 pixels for DenseNet201, MobileNetV2, and ResNet50, and 299×299 pixels for InceptionV3 and Xception). The pixel values are then normalized between 0 and 1 to ensure numerical stability and faster convergence during model inference. 

After preprocessing, we load the pre-trained CNN models, trained on ImageNet, and remove their final classification layers, retaining only the convolutional and pooling layers responsible for learning visual representations. The preprocessed images are then passed through the modified networks, and feature maps are extracted from the penultimate layer. These extracted feature vectors capture essential spatial and structural patterns, including textures, edges, shapes, and complex spatial structures, providing a compact yet informative summary of each CT scan. Finally, these feature representations are fed into conventional machine learning classifiers to diagnose brain strokes.  

\subsubsection{MobileNetV2}
MobileNetV2 is a lightweight deep neural network optimized for mobile and edge devices. It employs depthwise separable convolutions to reduce computational complexity and inverted residuals with linear bottlenecks to preserve representational power \cite{mnt1}\cite{mnt2}. This architecture is designed for efficient feature extraction while maintaining high performance.

\subsubsection{DensNet201}
DenseNet201 is designed to connect each layer to every other layer in a feedforward fashion \cite{dense1}. Unlike traditional architectures where layers connect only to subsequent layers, DenseNet enables feature reuse by concatenating feature maps from all preceding layers. This not only strengthens feature propagation but also reduces the number of parameters and mitigates vanishing gradients \cite{dense2}. 

\subsubsection{InceptionV3}
InceptionV3 is a convolutional neural network model designed to process images at multiple scales simultaneously using inception modules \cite{inception1}. These modules consist of parallel convolutional filters of different sizes (e.g., 1x1, 3x3, 5x5) that enable the model to capture both fine-grained and coarse features. It includes auxiliary classifiers and factorized convolutions to improve performance and reduce computational cost \cite{inception2}.

\subsubsection{ResNet50}
ResNet50 is a deep residual network with 50 layers known for introducing residual connections that bypass one or more layers \cite{rn1}. These shortcuts allow gradients to flow directly through the network, solving the vanishing gradient problem and enabling the training of very deep models \cite{rn2}. ResNet50 captures hierarchical spatial features effectively, making it suitable for image feature extraction \cite{rn3}.

\subsubsection{Xception}
Xception, short for "Extreme Inception," builds upon the Inception architecture by replacing standard convolutions with depthwise separable convolutions  \cite{xception1}. This decouples cross-channel and spatial correlations, significantly improving computational efficiency while maintaining model performance \cite{xception2}.

\subsection{Feature Engineering}
Feature engineering is the process of creating, transforming, or selecting relevant features from raw data to improve a model's performance \cite{featureengineering}. Feature selection is a subset of feature engineering that focuses on identifying the most relevant features for the model by eliminating redundant or irrelevant ones \cite{featureseleciton}. Feature reduction, on the other hand, reduces the dimensionality of the data while retaining essential information \cite{featurereduction}. Together, these processes enhance model efficiency, reduce overfitting, and improve interpretability. In this research we employed BFO as feature selection technique and PCA and LDA as feature reduction techniques. Bacterial Foraging Optimization (BFO) is a bio-inspired optimization algorithm based on the foraging behavior of bacteria \cite{bfo1}. It is used for feature selection by finding the optimal subset of features that maximize classification accuracy or minimize a predefined cost function \cite{bfo2}. BFO mimics bacterial processes like chemotaxis (movement toward nutrients or away from harmful substances), reproduction, elimination, and dispersal \cite{bfo3}. Principal Component Analysis (PCA) is an unsupervised dimensionality reduction technique used to transform a high-dimensional feature space into a lower-dimensional space by identifying the most significant components (principal components) \cite{pca1} \cite{pca2}. These components are linear combinations of the original features and are selected to maximize variance while minimizing redundancy. PCA assumes that most of the information in the data lies in the directions of maximum variance. PCA is particularly useful for reducing noise and computational complexity in high-dimensional datasets \cite{Sentiment}. Linear Discriminant Analysis (LDA) is a supervised feature reduction technique designed to maximize the separability between classes in the dataset  \cite{lda1}. Unlike PCA, which focuses on variance, LDA identifies the directions (linear discriminants) that maximize the ratio of between-class variance to within-class variance  \cite{lda2} \cite{lda3}. This ensures that the projected features enhance class discrimination.

\subsection{Classification Models}
In this research we utilized a combination of pre-trained deep learning models for feature extraction and machine learning classifiers for categorizing stroke into distinct classes. After the features were extracted from CT images and applying feature optimization, leveraging the representation capabilities of pre-trained models, and subsequently fed into various machine learning classifiers to perform classification. We utilized seven Machine learning classifiers. The selected classifiers are diverse in their methodologies, offering unique approaches to classification. Support Vector Classifier (SVC) is a supervised machine learning algorithm that seeks to find the optimal hyperplane separating data points of different classes. By using kernel functions, SVC can handle both linear and non-linear classification tasks effectively, making it a versatile choice for classification problems \cite{svm1, svm2}. Random Forest (RF) is an ensemble learning method that constructs multiple decision trees during training and aggregates their predictions for final output. RF is particularly robust against overfitting, handles both categorical and numerical data well, and provides feature importance scores, offering insights into the most influential features \cite{rf1}. Gaussian Naive Bayes (GNB) is based on Bayes theorem, this classifier assumes the independence of features within the dataset, which simplifies computation and makes it highly efficient for high-dimensional data. Despite its simplicity, GNB performs remarkably well in scenarios with normally distributed data \cite{nb1}. Decision Tree (DT) is a tree-structured algorithm that splits the dataset into branches based on feature values, producing a highly interpretable model. However, decision trees are prone to overfitting, especially when the tree becomes overly complex, which necessitates pruning or ensemble methods for better generalization \cite{dt1}. Extreme Gradient Boosting (XGBoost) is an advanced implementation of gradient boosting that improves classification performance by iteratively reducing errors from prior models. XGBoost is computationally efficient, handles missing data effectively, and supports regularization, making it a popular choice for structured datasets \cite{xgb1}. k-Nearest Neighbors (KNN) is a non-parametric algorithm that assigns class labels based on the majority vote of the nearest neighbors. While simple to implement, KNN can be computationally intensive for large datasets and is sensitive to the choice of distance metrics \cite{knn1} \cite{maternal}. Logistic Regression (LR) is a statistical model that estimates the probability of class membership using the logistic function. LR is effective for both binary and multi-class classification tasks, particularly when the relationship between features and the target variable is approximately linear \cite{lr1}. Each of these classifiers, combined with the extracted features and optimized through the selected techniques, is evaluated for its ability to maximize accuracy and generalization.

\section{Experimental Setup and Implementation}

\subsection{Environment}
The experimental study was conducted using a Windows 10 operating system on a ×64 processor, intel(R) core i5 8th Gen CPU running at 1.60GHz to 3.71GHz with 8 GB of DDR4 RAM. Anaconda Navigator was used with the Jupyter Notebook for image processing applications. 

The proposed model was trained on a three-class image dataset comprising CT scan images of Normal, Hemorrhagic, and Ischemic strokes. The preprocessed images were resized to standard dimensions of 224×224 pixels for most CNN models, except for InceptionV3 and Xception, which required image dimensions of 299×299 pixels. The model training employed 10-fold cross-validation for robust evaluation. Feature optimization techniques, including Bacterial Foraging Optimization (BFO), Principal Component Analysis (PCA), and Linear Discriminant Analysis (LDA), were used to refine the extracted features. Transfer learning was applied to accelerate training by leveraging pre-trained CNN models. This approach significantly reduced the training time. Each CNN model took approximately 2 hours to extract features from the image dataset. After applying feature optimization and training the classifiers, the process took nearly 3 hours for each feature extraction method. Since five different feature extraction techniques were employed the total training time was approximately 15 hours. Testing the trained models required approximately only 30 minutes, showcasing the efficiency of the proposed method.

\subsection{Parameters}
This section presents the essential details for five pre-trained deep learning models, including the number of parameters, Multiply-Accumulate operations (MACs), and Floating Point Operations (FLOPs). These metrics are crucial for understanding the computational efficiency and resource requirements of the models. 

ResNet50 and DenseNet201 have relatively high parameter counts of 25.557M and 20.014M, respectively, which suggests higher memory requirements. In contrast, MobileNetV2 has significantly fewer parameters, with only 3.505M, making it a more lightweight model. The MACs values, which indicate the total operations required for inference, vary across the architectures. MobileNetV2 requires 327.487M MACs, whereas ResNet50 and InceptionV3 require 4.134G and 5.749G MACs, respectively. Additionally, FLOPs, which measure the number of floating-point calculations, show that MobileNetV2 has the lowest computational cost at 654.973M, while InceptionV3 has the highest at 11.498G. These values highlight the trade-offs between model complexity and computational efficiency, influencing model selection based on available hardware resources. The detailed resource specifications of the experimental methods are displayed in Table \ref{tab:pt}.

\begin{table}[htbp]
\centering
\scriptsize
\setlength{\tabcolsep}{15pt}
\renewcommand{\arraystretch}{1.4}
\caption{Parameters and Resource Requirement of Experimental methods}
\begin{tabular}{llll}
\hline
\textbf{Model} & \textbf{Parameter} & \textbf{MACS} & \textbf{FPO} \\ \hline
ResNet50       & 25.557M            & 4.134G                               & 8.267G                             \\
DenseNet201    & 20.014M            & 4.390G                               & 8.781G                             \\
MobileNetV2    & 3.505M             & 327.487M                             & 654.973M                           \\
InceptionV3    & 23.835M            & 5.749G                               & 11.498G                            \\
Xeption        & 22.855M            & 4.146G                               & 8.292G                             \\ \hline
\end{tabular}
\label{tab:pt}
\end{table}

\subsection{Implementation Details}
The proposed brain stroke diagnosis model begins with collecting and preparing datasets, Dataset-1 and Dataset-2, which are then manually curated to ensure data quality. This includes removing irrelevant and duplicate entries and incorporating additional data from the local clinic to enhance the dataset's diversity and representativeness. The preprocessing stage involves resizing each image to consistent dimensions of 224×224 pixels for most CNN architectures and 299×299 pixels for Inception V3 and Xception. Rescaling is performed to normalize the pixel values to a range of [0, 1], ensuring compatibility with the input requirements of the candidate models. Following preprocessing, feature extraction is conducted using five pre-trained convolutional neural network (CNN) architectures: DenseNet201, InceptionV3, Xception, ResNet50, and MobileNet V2. The classification layers of these CNNs are removed, allowing the penultimate layer's feature maps to be extracted. These extracted features are saved as high-dimensional vectors in a CSV format for further processing. The feature vectors are then subjected to either feature selection and dimensionality reduction techniques or directly used without feature selection. Feature reduction methods include Bacterial Foraging Optimization (BFO), A bio-inspired optimization technique for selecting the most relevant features. Principal Component Analysis (PCA) is a statistical method to reduce dimensionality while preserving variance. Linear Discriminant Analysis (LDA) is a technique to maximize class separability by projecting data into a lower-dimensional space. The processed features are subsequently fed into a set of traditional machine learning classifiers: Support Vector Classifier (SVC), Random Forest (RF), Gaussian Naive Bayes (GNB), Decision Tree (DT), Extreme Gradient Boosting (XGB), K-Nearest Neighbors (KNN), and Logistic Regression (LR). Each classifier is trained and evaluated to determine its performance in predicting stroke categories. Finally, the best-performing model is selected based on its performance metrics and is utilized to classify outcomes into three categories: hemorrhagic stroke, ischemic stroke, and normal condition. This end-to-end approach ensures a robust and accurate diagnosis system suitable for real-world clinical applications. The algorithm \ref{alg:brainstroke} shows the implementation procedure of the proposed brain stroke detection method.
\begin{algorithm}[]
\scriptsize
\caption{Algorithm for Brain Stroke Detection}
\label{alg:brainstroke}

\textbf{Initialize:} $DS_1$, $DS_2$, $PP$, $DA$, $CNN$, $FE$, $FEG$, $ML$, $EVAL$, $TC$, $SC$ \tcp*[r]{Initialize components}

\textbf{Step 1: Data Collection} \;
$DS_1 \gets \text{Brain Stroke CT Image Dataset}$ \;
$DS_2 \gets \text{Brain Stroke Prediction CT Scan Image Dataset}$ \;

\textbf{Step 2: Dataset Preparation} \;
Remove 424 images from $DS_1$ \tcp*[r]{Selectively remove Normal images} 
$DS' \gets DS_1[\text{Selected Normal}] \cup DS_2[\text{Ischemic, Hemorrhagic}]$ \tcp*[r]{Create initial dataset} 
Add 177 Normal images from Local Hospital to $DS'$ \tcp*[r]{Balance Normal data in the final dataset} 
$DS \gets DS'$ \tcp*[r]{Final dataset} 

\textbf{Step 3: Data Preprocessing} \;
$PP \gets [\text{Resizing, Rescaling}]$ \;
Apply $PP$ on all images in $DS$ \;

\textbf{Step 4: Data Augmentation} \;

$DA \gets [\text{Random Flipping, Rotation, Scaling, and Cropping}]$ \;

Apply $DA$ on $DS$ \tcp*[r]{Generate Additional Samples} 

\textbf{Step 5: Feature Extraction Using CNN} \;

$CNN \gets [\text{MobileNetV2, ResNet50, DenseNet201, 
InceptionV3, Xception}]$ \;

\For{$i \gets 1$ \KwTo $|CNN|$}{
    $FV \gets FE(CNN[i], DS[\text{Images}])$ \tcp*[r]{Extract Features} 
}

\textbf{Step 6: Feature Optimization} \;
$FEG \gets [\text{None, BFO, PCA, LDA}]$ \tcp*[r]{Feature Optimization Techniques} 

\For{$j \gets 1$ \KwTo $|FEG|$}{
    $FV' \gets \text{Apply } FEG[j] \text{ on } FV$ \tcp*[r]{Apply Feature Optimization} 
}

\textbf{Step 7: Classification Using ML Models} \;

$ML \gets [\text{SVC, RF, DT, LR, GNB, XGB, KNN}]$ \tcp*[r]{ML Classifiers} 

\For{$k \gets 1$ \KwTo $|ML|$}{
    \tcp{Train-Test-Val Split}
    $X_T, Y_T, x_t, y_t \gets \text{Split Data}(FV', DS[\text{Labels}])$ \;
    
    \tcp{Train ML Classifier}
    $ML[k].\text{fit}(X_T, Y_T)$ \;

    \tcp{Test ML Classifier}
    $y_{\text{pred}} \gets ML[k].\text{predict}(x_t)$ \;

    \tcp{Evaluation Metrics}
    $EVAL \gets [\text{Accuracy, Precision, Recall, F1-Score, AUC}]$ \;
    Compute $EVAL$ for $ML[k]$ \;
}

\textbf{Step 8: Performance Comparison} \;

Compute computational cost:  
$TC \gets \text{Time Complexity}$ \;
$SC \gets \text{Space Complexity}$ \;

\textbf{Step 9: Final Model Selection} \;

Select the best model based on $EVAL$, $TC$, and $SC$ \;

\textbf{Deinitialize:} $DS_1$, $DS_2$, $PP$, $DA$, $CNN$, $FE$, $FEG$, $ML$, $EVAL$, $TC$, $SC$ \tcp*[r]{Shutdown}
\end{algorithm}

\subsection{Evaluation Matrices}
In this research, The major evaluation metrics accuracy, recall, precision, and F1-score are used to assess the experimental model. The following are these metrics definitions and equations: \vspace{4mm}

\begin{itemize}
  
\item Accuracy is defined as the proportion of correctly classified cases to all instances.

\begin{equation}
\scriptsize
\text{Accuracy} = \left( \frac{TN + TP}{(FN + TN) + (TP + FP)} \right) \times 100
\end{equation} \vspace{2mm}

\item Precision computes the proportion of true positives among all samples classified as positive, indicating how successfully the classifier can identify each class.
\begin{equation}
\scriptsize
\text{Precision} = \left( \frac{TP}{FP + TP} \right) \times 100
\end{equation}

\item Recall measures how well the model detects positive results; it is expressed as the ratio of true positives to the sum of false negatives and true positives.
\begin{equation}
\scriptsize
\text{Recall} = \left( \frac{TP}{FN + TP} \right) \times 100
\end{equation}

\item The F1 Score is a statistic that assesses the overall performance using the harmonic mean of recall and accuracy.
\begin{equation}
\scriptsize
\text{F1 Score} = 2 \times \frac{\text{Recall} \times \text{Precision}}{\text{Recall} + \text{Precision}} \times 100
\end{equation}

\item The Receiver Operating Characteristic curve, or ROC curve, is a visual representation of a classifier's performance over all classification levels. The True Positive Rate (TPR), which is defined as:
\begin{equation}
\scriptsize
\text{TPR} = \frac{\text{TP}}{\text{TP} + \text{FN}}
\end{equation}

and False Positive Rate (FPR) is defined as:

\begin{equation}
\scriptsize
\text{FPR} = \frac{\text{FP}}{\text{FP} + \text{TN}}
\end{equation}
\end{itemize}

\section{Results and Discussion}

\subsection{Result Analysis}
In this section, we present the performance evaluation of various machine learning models and techniques employed in our study. The aim is to analyze the effectiveness of each approach in terms of multiple performance matrices. By examining the learning curves, confusion matrix, ROC curve, and tables, we compare the strengths and limitations of each model to determine the most suitable method for the brain stroke diagnosis. The results provide insights into the impact of optimization techniques, feature reduction methods, and classifier combinations on model performance.

Table \ref{tab:r1} illustrates the classification performance (without using any optimization) of various pre-trained models DenseNet201, InceptionV3, MobileNetV2, ResNet50, and Xception combined with different classifiers SVC, RF, GNB, DT, XGB, KNN, and LR, across four evaluation metrics: accuracy, precision, recall, and F1-score. Among the combinations, MobileNetV2 with the XGB classifier achieved the highest overall performance, with an accuracy of 89.21\%, precision of 89.39\%, recall of 88.91\%, and F1-score of 89.11\%. In contrast, Xception paired with GNB and DT yielded the lowest accuracy at 23.86\%. Across models, MobileNetV2 consistently demonstrated superior results, particularly when paired with ensemble or nearest neighbor-based classifiers, while other models showed comparatively lower classification performance. The results underscore the effectiveness of lightweight architectures such as MobileNetV2 when coupled with ML classifiers for brain stroke diagnoses.

\begin{table}[htbp]
\caption{Classification performance of different pre-trained models and classifiers (No optimization) for brain stroke detection} \vspace{1mm}
\centering
\scriptsize
\label{tab:r1}
\renewcommand{\arraystretch}{1.3}
\resizebox{\columnwidth}{!}{%
\begin{tabular}{|c|c|c|c|c|c|}
\hline
\textbf{Pre-trained} & \textbf{Classifier} & \textbf{Accuracy} & \textbf{Precision} & \textbf{Recall} & \textbf{F1-Score} \\
\textbf{Model} & & \textbf{(\%)} & \textbf{(\%)} & \textbf{(\%)} & \textbf{(\%)} \\
\hline
\multirow{7}{*}{DenseNet201} & SVC & 44.07 & 32.35 & 38.54 & 31.03 \\
 & RF & 37.42 & 38.68 & 37.59 & 37.98 \\
 & GNB & 42.00 & 41.02 & 43.22 & 41.30 \\
 & DT & 37.42 & 38.68 & 37.59 & 42.00 \\
 & XGB & 39.71 & 43.55 & 40.15 & 41.52 \\
 & KNN & 32.64 & 36.73 & 32.30 & 32.95 \\
 & LR & 39.92 & 37.02 & 38.36 & 36.98 \\
\hline
\multirow{7}{*}{InceptionV3} & SVC & 45.64 & 42.92 & 44.20 & 42.19 \\
 & RF & 34.02 & 33.01 & 34.15 & 33.38 \\
 & GNB & 38.59 & 37.71 & 39.29 & 37.23 \\
 & DT & 25.73 & 30.36 & 26.88 & 28.11 \\
 & XGB & 40.25 & 43.14 & 40.64 & 41.66 \\
 & KNN & 39.21 & 40.72 & 38.98 & 39.59 \\
 & LR & 36.93 & 38.27 & 37.41 & 37.81 \\
\hline
\multirow{7}{*}{MobileNetV2} 
 & SVC & 86.51 & 86.76 & 86.02 & 86.30 \\
 & RF & 84.44 & 85.50 & 83.07 & 83.59 \\
 & GNB & 74.90 & 75.18 & 75.36 & 75.22 \\
 & DT & 78.22 & 78.34 & 78.68 & 78.46 \\
 & \textbf{XGB} & \textbf{89.21} & \textbf{89.39} & \textbf{88.91} & \textbf{89.11} \\
 & KNN & 87.55 & 87.54 & 87.81 & 87.65 \\
 & LR & 86.72 & 86.73 & 86.82 & 86.77 \\
\hline
\multirow{7}{*}{ResNet50} & SVC & 50.21 & 45.08 & 47.15 & 42.43 \\
 & RF & 33.82 & 33.45 & 34.53 & 33.93 \\
 & GNB & 39.21 & 37.47 & 39.49 & 38.15 \\
 & DT & 28.42 & 30.76 & 29.88 & 30.13 \\
 & XGB & 39.63 & 42.86 & 39.72 & 40.87 \\
 & KNN & 37.97 & 43.76 & 38.23 & 39.45 \\
 & LR & 40.04 & 43.78 & 41.19 & 42.19 \\
\hline
\multirow{7}{*}{Xception} & SVC & 44.40 & 44.18 & 43.07 & 42.52 \\
 & RF & 35.89 & 36.77 & 36.45 & 36.54 \\
 & GNB & 23.86 & 26.82 & 24.99 & 25.59 \\
 & DT & 23.86 & 26.82 & 24.99 & 25.59 \\
 & XGB & 41.29 & 41.57 & 42.64 & 41.03 \\
 & KNN & 37.14 & 42.49 & 37.57 & 39.44 \\
 & LR & 39.21 & 37.47 & 39.49 & 38.15 \\
\hline
\end{tabular}
}
\end{table}

Table \ref{tab:r2} presents the performance analysis of various pre-trained models DenseNet201, InceptionV3, MobileNetV2, ResNet50, and Xception integrated with different classifiers SVC, RF, GNB, DT, XGB, KNN, and LR for brain stroke detection, incorporating Bacterial Foraging Optimization (BFO). MobileNetV2 combined with the KNN classifier demonstrates the best performance, achieving an accuracy of 90.66\%, precision of 90.87\%, recall of 90.52\%, and F1-score of 90.67\%. Conversely, Xception with the GNB classifier shows the lowest accuracy at 25.31\%. MobileNetV2 emerged as the most effective model, particularly when paired with KNN and XGB classifiers, outperforming other combinations across all metrics. The table highlights the enhancement in classification performance due to the optimization technique, particularly for lightweight architectures like MobileNetV2, showcasing its potential for brain stroke detection.

\begin{table}[htbp]
\centering
\scriptsize
\caption{Classification performance of different pre-trained models and classifiers with Bacterial Foraging optimization for brain stroke detection}\vspace{2mm}
\label{tab:r2}
\renewcommand{\arraystretch}{1.3}
\resizebox{\columnwidth}{!}{%
\begin{tabular}{|c|c|c|c|c|c|}
\hline
\textbf{Pre-trained } & \textbf{Classifier} & \textbf{Accuracy} & \textbf{Precision} & \textbf{Recall} & \textbf{F1-Score} \\
\textbf{model} & & \textbf{(\%)} & \textbf{(\%)} & \textbf{(\%)} & \textbf{(\%)} \\
\hline
\multirow{7}{*}{DenseNet201} & SVC & 42.62 & 31.91 & 37.66 & 28.14 \\
 & RF & 37.66 & 28.14 & 42.20 & 41.10 \\
 & GNB & 41.62 & 40.57 & 29.11 & 32.13 \\
 & DT & 29.86 & 30.55 & 39.29 & 39.48 \\
 & XGB & 44.28 & 46.68 & 43.88 & 44.63 \\
 & KNN & 34.10 & 37.62 & 33.68 & 34.33 \\
 & LR & 40.33 & 34.18 & 37.47 & 32.90 \\
\hline
\multirow{7}{*}{InceptionV3} & SVC & 41.70 & 39.00 & 40.00 & 37.92 \\ 
 & RF & 35.06 & 33.86 & 34.22 & 33.30 \\
 & GNB & 28.01 & 30.29 & 28.40 & 29.15 \\
 & DT & 36.51 & 35.91 & 37.21 & 35.67 \\
 & XGB & 38.38 & 39.79 & 37.75 & 38.11 \\
 & KNN & 41.49 & 44.69 & 41.75 & 42.54 \\
 & LR & 36.10 & 36.92 & 36.21 & 36.51 \\
\hline
\multirow{7}{*}{MobileNetV2} & SVC & 81.95 & 83.35 & 80.71 & 80.91 \\
 & RF & 80.08 & 82.79 & 78.22 & 78.01 \\
 & GNB & 71.99 & 72.07 & 72.17 & 72.12 \\
 & DT & 72.20 & 72.48 & 72.65 & 72.54 \\
 & XGB & 87.14 & 87.89 & 86.50 & 86.87 \\
 & \textbf{KNN} & \textbf{90.66} & \textbf{90.87} & \textbf{90.52} & \textbf{90.67} \\
 & LR & 81.74 & 82.06 & 81.27 & 81.47 \\
\hline
\multirow{7}{*}{ResNet50} & SVC & 45.02 & 42.43 & 42.32 & 37.76 \\

 & RF & 38.80 & 37.78 & 38.81 & 37.93 \\
 & GNB & 27.18 & 28.80 & 28.05 & 28.38 \\
 & DT & 37.97 & 37.35 & 38.08 & 37.61 \\
 & XGB & 37.34 & 40.38 & 37.01 & 37.91 \\
 & KNN & 40.04 & 44.73 & 39.98 & 41.16 \\
 & LR & 42.53 & 43.03 & 42.61 & 42.71 \\
\hline
\multirow{7}{*}{Xception} & SVC & 41.70 & 38.60 & 39.57 & 36.54 \\ 
 & RF & 35.27 & 33.70 & 35.08 & 33.96 \\
 & GNB & 25.31 & 26.51 & 25.91 & 26.15 \\
 & DT & 37.76 & 37.72 & 38.63 & 37.91 \\
 & XGB & 35.06 & 36.67 & 34.59 & 35.05 \\
 & KNN & 43.15 & 46.02 & 43.58 & 44.42 \\
 & LR & 39.00 & 39.93 & 38.90 & 39.16 \\
\hline
\end{tabular}
}
\end{table}

Table \ref{tab:r3} summarizes the performance of various classifiers integrated with different pre-trained models optimized using PCA across four metrics: accuracy, precision, recall, and F1-score. For each pre-trained model DenseNet201, InceptionV3, MobileNetV2, ResNet50, and Xception, multiple classifiers, including SVC, RF, GNB, DT, XGB, KNN, and LR, were evaluated. MobileNetV2 combined with the SVC classifier achieved the highest performance, with an accuracy of 86.03\%, precision of 86.79\%, recall of 85.67\%, and an F1-score of 85.95\%. In contrast, DenseNet201 with GNB showed the lowest performance across metrics, with accuracy at 26.49\%, precision at 28.91\%, recall at 26.78\%, and an F1-score of 27.51\%. These results highlight the significant variability in classifier performance depending on the combination of the pre-trained model and the algorithm used.

\begin{table}[htbp]
\centering
\scriptsize
\renewcommand{\arraystretch}{1.3}
\caption{Classification performance of different pre-trained models and classifiers with Principle Component Analysis for brain stroke detection} \vspace{2 mm}
\label{tab:r3}
\resizebox{\columnwidth}{!}{%
\begin{tabular}{|c|c|c|c|c|c|}
\hline
\textbf{Pre-trained} & \textbf{Classifier} & \textbf{Accuracy} & \textbf{Precision} & \textbf{Recall} & \textbf{F1-Score} \\
\textbf{model} & & \textbf{(\%)} & \textbf{(\%)} & \textbf{(\%)} & \textbf{(\%)} \\
\hline
\multirow{7}{*}{DenseNet201} & SVC & 50.90 & 52.98 & 49.29 & 50.05 \\

 & RF & 37.03 & 38.43 & 35.67 & 36.22 \\
 & GNB & 26.49 & 28.91 & 26.78 & 27.51 \\
 & DT & 39.11 & 38.32 & 39.69 & 38.37 \\
 & XGB & 37.86 & 41.75 & 37.27 & 38.81 \\
 & KNN & 51.46 & 56.78 & 51.68 & 52.89 \\
 & LR & 47.02 & 48.49 & 46.79 & 47.49 \\
\hline
\multirow{7}{*}{InceptionV3} & SVC & 41.77 & 40.78 & 41.30 & 39.66 \\ 
 & RF & 28.22 & 24.55 & 29.07 & 26.39 \\
 & GNB & 28.35 & 30.08 & 29.24 & 29.59 \\
 & DT & 32.37 & 33.54 & 35.89 & 27.64 \\
 & XGB & 32.23 & 31.14 & 33.21 & 31.89 \\
 & KNN & 43.29 & 45.44 & 43.55 & 44.24 \\
 & LR & 35.13 & 36.60 & 36.03 & 36.29 \\
\hline
\multirow{7}{*}{MobileNetV2} & \textbf{SVC} & \textbf{86.03} & \textbf{86.79} & \textbf{85.67} & \textbf{85.95} \\ 
 & RF & 78.98 & 82.66 & 77.40 & 76.75 \\
 & GNB & 71.51 & 72.67 & 72.75 & 72.66 \\
 & DT & 81.60 & 81.96 & 81.22 & 81.24 \\
 & XGB & 83.96 & 85.71 & 83.41 & 83.73 \\
 & KNN & 85.93 & 85.56 & 85.29 & 85.40 \\
 & LR & 79.81 & 80.58 & 80.00 & 80.16 \\
\hline
\multirow{7}{*}{ResNet50} & SVC & 42.46 & 43.10 & 40.69 & 40.29 \\
 
 & RF & 33.33 & 32.79 & 32.66 & 32.31 \\
 & GNB & 31.95 & 34.15 & 34.68 & 28.68 \\
 & DT & 31.95 & 34.15 & 34.68 & 28.68 \\
 & XGB & 30.57 & 31.18 & 30.31 & 30.58 \\
 & KNN & 43.71 & 47.17 & 43.63 & 44.01 \\
 & LR & 37.62 & 41.53 & 38.17 & 39.53 \\
\hline
\multirow{7}{*}{Xception} & SVC & 44.12 & 44.84 & 43.02 & 43.34 \\
 & RF & 30.01 & 27.50 & 30.58 & 28.77 \\
 & GNB & 27.25 & 28.32 & 28.77 & 28.27 \\
 & DT & 28.77 & 27.77 & 32.72 & 25.17 \\
 & XGB & 33.47 & 33.05 & 34.31 & 33.63 \\
 & KNN & 41.36 & 44.37 & 42.04 & 42.52 \\
 & LR & 37.07 & 38.85 & 38.25 & 38.50 \\
\hline
\end{tabular}
}
\end{table}

Table \ref{tab:r4} illustrates the performance of different pre-trained models, DenseNet201, InceptionV3, MobileNetV2, ResNet50, and Xception, combined with classifiers, SVC, RF, GNB, DT, XGB, KNN, and LR, using Linear Discriminant Analysis (LDA) optimization for classification. Among the combinations, MobileNetV2 paired with the SVC classifier achieves the highest accuracy of 97.93\%, precision of 98.02\%, recall of 97.80\%, and F1-score of 97.90\%, showcasing the most robust performance across all metrics. On the other hand, Xception with the GNB classifier records the lowest accuracy of 26\%. MobileNetV2 consistently outperforms other models across most classifiers, highlighting its adaptability and effectiveness when optimized with LDA. These results underline the utility of LDA optimization in enhancing the discriminative capability of pre-trained models for classification tasks. \vspace{4mm}

After analyzing all models and optimization techniques, we found that MobileNetV2 consistently outperformed all other deep learning models across all optimization techniques. Among these combinations, MobileNetV2 with LDA and SVC achieved the best overall performance. This combination demonstrated superior accuracy and balanced classification metrics, making it the most effective approach for brain stroke detection in our study.

\begin{table}[]
\centering
\scriptsize
\caption{Classification performance of different pre-trained models and classifiers with Linear Discriminant Analysis for brain stroke detection} \vspace{2 mm}
\renewcommand{\arraystretch}{1.3}
\label{tab:r4}
\resizebox{\columnwidth}{!}{%
\begin{tabular}{| c | c | c | c | c | c |}
\hline
\textbf{Pre-trained} & \textbf{Classifier} & \textbf{Accuracy} & \textbf{Precision} & \textbf{Recall} & \textbf{F1-Score} \\
\textbf{model} & & \textbf{(\%)} & \textbf{(\%)} & \textbf{(\%)} & \textbf{(\%)} \\
\hline
\multirow{7}{*}{DenseNet201} & SVC & 47.43 & 50.67 & 46.91 & 48.08 \\

& RF & 37.73 & 38.41 & 37.91 & 38.06 \\
& GNB & 26.35 & 29.70 & 27.05 & 28.04 \\
& DT & 39.39 & 39.58 & 40.40 & 39.22 \\
& XGB & 41.75 & 45.40 & 41.80 & 43.17 \\
& KNN & 47.16 & 53.31 & 47.92 & 49.12 \\
& LR & 39.39 & 42.59 & 40.45 & 41.40 \\
\hline
\multirow{7}{*}{InceptionV3} & SVC & 39.83 & 41.65 & 38.76 & 39.83 \\

& RF & 34.44 & 34.05 & 34.21 & 34.13 \\
& GNB & 28.08 & 31.34 & 29.12 & 29.65 \\
& DT & 37.48 & 36.89 & 37.83 & 35.98 \\
& XGB & 37.07 & 41.32 & 36.83 & 38.45 \\
& KNN & 40.11 & 43.44 & 41.10 & 40.94 \\
& LR & 33.75 & 36.57 & 33.61 & 34.75 \\
\hline
\multirow{7}{*}{MobileNetV2} & \textbf{SVC} & \textbf{97.93} & \textbf{98.02} & \textbf{97.80} & \textbf{97.90} \\

& RF & 81.88 & 81.88 & 84.84 & 79.93 \\
& GNB & 79.69 & 77.18 & 76.60 & 76.52 \\
& DT & 76.52 & 73.58 & 73.14 & 73.16 \\
& XGB & 73.14 & 90.73 & 90.16 & 90.42 \\
& KNN & 87.69 & 87.68 & 87.16 & 87.33 \\
& LR & 84.79 & 84.58 & 84.40 & 84.48 \\
\hline
\multirow{7}{*}{ResNet50} & SVC & 40.25 & 40.25 & 42.39 & 39.87 \\

& RF & 39.30 & 36.51 & 38.20 & 36.35 \\
& GNB & 36.04 & 28.63 & 29.28 & 29.77 \\
& DT & 39.14 & 39.64 & 40.11 & 39.10 \\
& XGB & 37.34 & 40.46 & 37.58 & 38.37 \\
& KNN & 39.97 & 45.32 & 40.09 & 40.90 \\
& LR & 38.17 & 41.30 & 38.56 & 39.55 \\
\hline
\multirow{7}{*}{Xception} & SVC & 39.14 & 39.19 & 39.15 & 36.99 \\

& RF & 33.61 & 33.46 & 34.22 & 32.98 \\
& GNB & 26.00 & 28.35 & 26.73 & 27.41 \\
& DT & 39.42 & 39.04 & 40.64 & 39.19 \\
& XGB & 35.27 & 38.24 & 35.82 & 36.15 \\
& KNN & 41.77 & 46.37 & 41.81 & 43.07 \\
& LR & 35.82 & 40.09 & 36.59 & 37.84 \\
\hline
\end{tabular}
}
\end{table}

\begin{figure*}[htbp]
    \centering
 \subfigure[]{\label{fig:aa}\includegraphics[height=3.1cm, width=4.52cm]{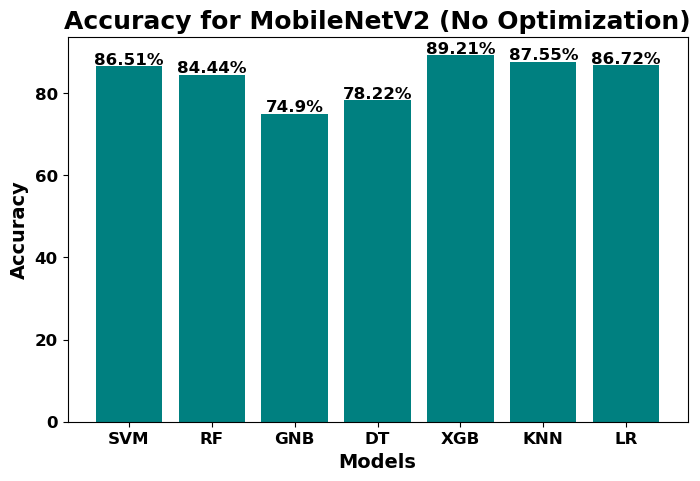}}
 \subfigure[]{\label{fig:bb}\includegraphics[height=3.1cm, width=4.52cm]{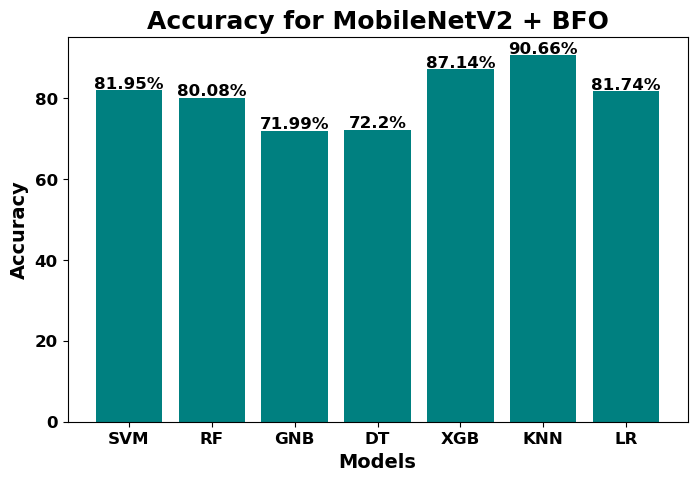}}
 \subfigure[]{\label{fig:cc}\includegraphics[height=3.1cm, width=4.52cm]{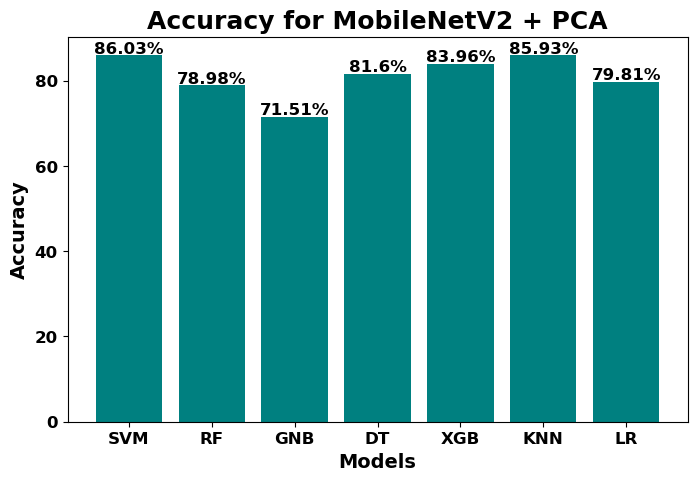}}
 \subfigure[]{\label{fig:dd}\includegraphics[height=3.1cm, width=4.52cm]{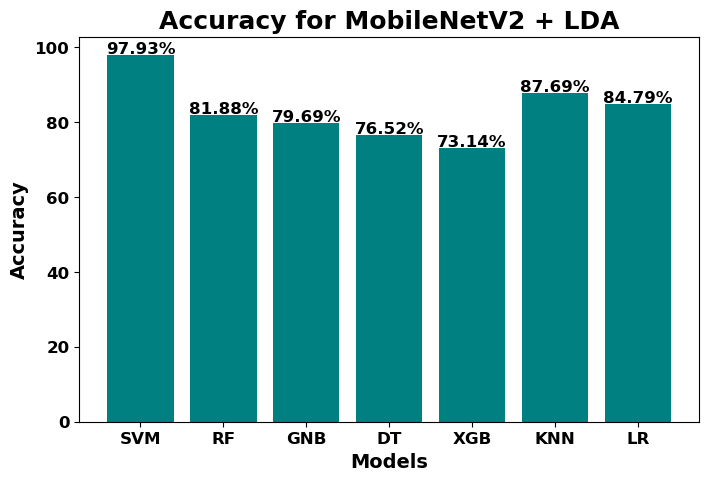}}
   \vspace{-0.4cm}
    \caption{Bar Charts compare the classification performance of different feature selection techniques applied with MobileNetV2 and different classifiers: (a) No Feature Selection, (b) Bacterial Foraging Optimization  (BFO), (c) Principal Component Analysis (PCA), and (d) Linear Discriminant Analysis (LDA). Each feature selection technique shows the comparison of accuracy across seven different classifiers.}
    \label{fig:ac}
\end{figure*}

\begin{figure*}[]
    \centering
     \subfigure[]{\label{fig:ee}\includegraphics[height=3.1cm, width=4.52cm]{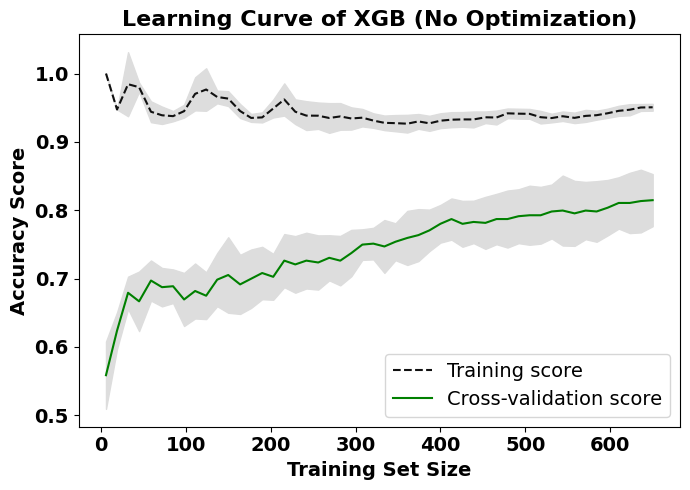}}
    \subfigure[]{\label{fig:ff}\includegraphics[height=3.1cm, width=4.52cm]{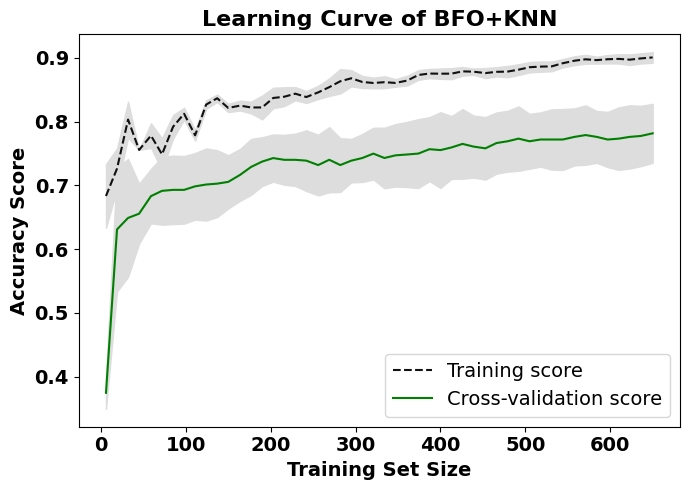}}
    \subfigure[]{\label{fig:gg}\includegraphics[height=3.1cm, width=4.52cm]{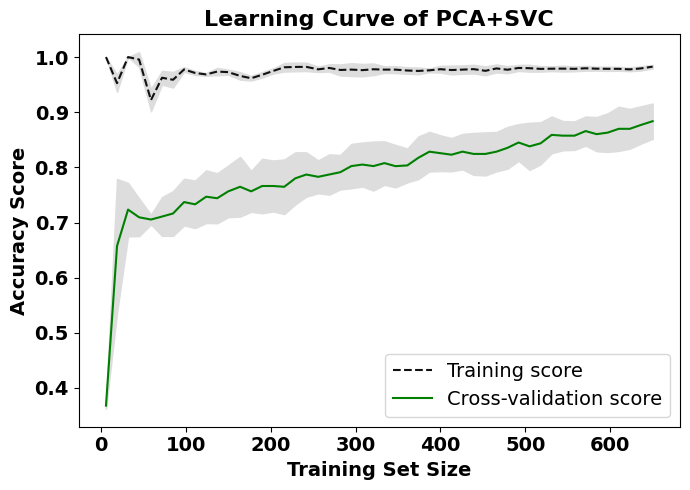}}
    \subfigure[]{\label{fig:hh}\includegraphics[height=3.1cm, width=4.52cm]{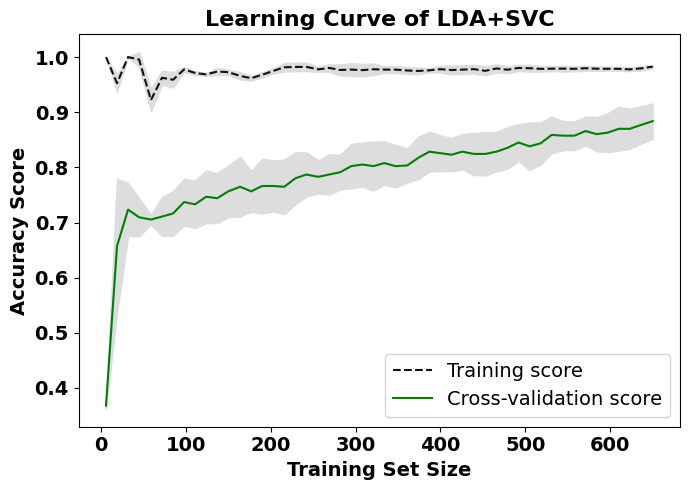}}
    \vspace{-0.4cm}
    \caption{Learning curves comparing the performance of different feature optimization techniques employed with MobileNetV2 in terms of accuracy score with varying training set sizes. (a) Extreme Gradient Boosting with No Feature Optimization, (b) Bacterial Foraging Optimization with k-Nearest Neighbors, (c) Principal Component Analysis with Support Vector Classifier, (d) Linear Discriminant Analysis and Support Vector Classifier. \vspace{-2mm}}
    \label{fig:lc}
\end{figure*}

Table \ref{tab:r5} presents the performance evaluation of the proposed MobileNetV2 + LDA + SVC model using 10-fold cross-validation across four metrics: accuracy, precision, recall, and F1-score. The results show consistent and high performance across all folds, with accuracy ranging from 95.15\% to 100\%, precision from 94.67\% to 100\%, recall from 95.03\% to 100\%, and F1-score from 94.97\% to 100\%. The mean values for the metrics are 97.93\% accuracy, 98.02\% precision, 97.80\% recall, and 97.90\% F1-score, indicating the model's reliability and robustness in classification tasks. \vspace{3mm}

\begin{table}[]
\centering
\scriptsize
\caption{10 Fold Cross Validation For MobileNetV2+LDA+SVC)}
\label{tab:r5}
\renewcommand{\arraystretch}{.9}
\resizebox{\columnwidth}{!}{
\begin{tabular}{| c | c | c | c | c |}
\hline
\textbf{Number of} & \textbf{Accuracy} & \textbf{Precision} & \textbf{Recall} & \textbf{F1-Score} \\
 \textbf{Fold} & \textbf{(\%)} & \textbf{(\%)} & \textbf{(\%)} & \textbf{(\%)} \\
\hline
Fold 1 & 95.64 & 95.51 & 95.51 & 95.65 \\
\hline
Fold 2 & 97.09 & 97.53 & 96.96 & 97.09 \\
\hline
Fold 3 & 99.98 & 99.79 & 99.85 & 99.83 \\
\hline
Fold 4 & 100 & 100 & 100 & 100 \\
\hline
Fold 5 & 100 & 100 & 100 & 100 \\
\hline
Fold 6 & 98.53 & 98.96 & 98.40 & 98.60 \\
\hline
Fold 7 & 95.15 & 94.67 & 95.03 & 94.97 \\
\hline
Fold 8 & 96.03 & 97.03 & 95.91 & 96.27 \\
\hline
Fold 9 & 96.52 & 97.11 & 96.39 & 96.54 \\
\hline
Fold 10 & 97.49 & 97.11 & 97.36 & 97.33 \\
\hline
\textbf{Mean} & \textbf{97.93} & \textbf{98.02} & \textbf{97.80} & \textbf{97.90} \\
\hline
\end{tabular}
}
\end{table}

\begin{figure}[htbp]
	\centering 
	\fbox{\includegraphics[width=0.47\textwidth, angle=0]{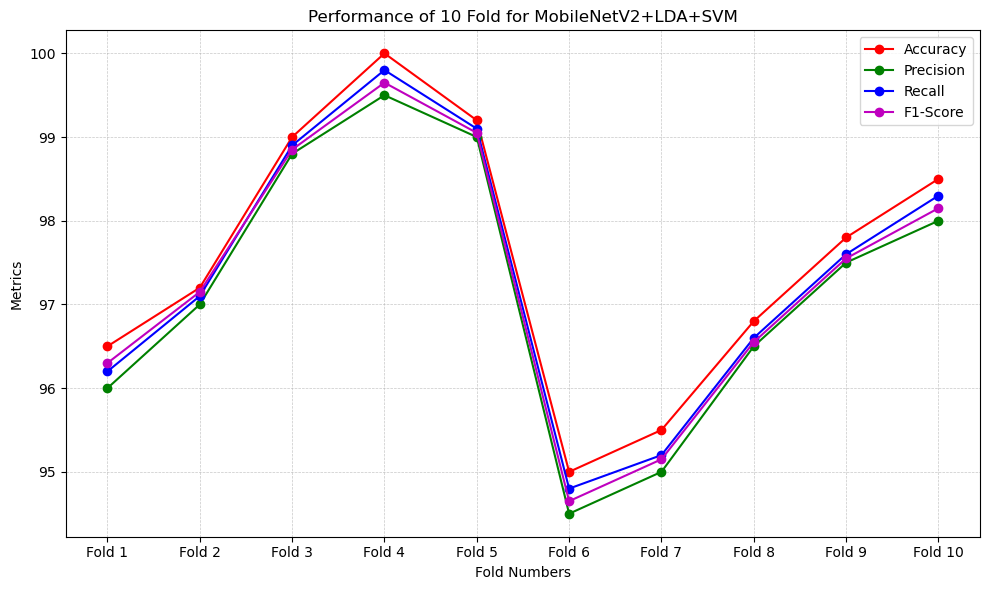}}	
		\caption{{10 Fold Cross Validation For MobileNetV2+LDA+SVC} }
	\label{fig:cva}
\end{figure}

The Fig. \ref{fig:cva} illustrates the 10-fold cross-validation performance of MobileNetV2 with Linear Discriminant Analysis (LDA) and Support Vector Classifier (SVC), showing accuracy, precision, recall, and F1-score across different folds. After comparing various models, classifiers, and feature optimization combinations, we identified this approach as the best-performing one. The model demonstrates strong and stable performance, with accuracy peaking close to 100\% in Fold 4, while Fold 6 shows a slight dip below 95\% across all metrics. Despite minor fluctuations, the overall trend remains consistently high, confirming the reliability and effectiveness of this method.

Fig. \ref{fig:ac} presents a comparison of classification accuracies across different machine learning models SVC, RF, GNB, DT, XGB, KNN, and LR integrated with MobileNetV2 under four feature selection techniques: no feature selection, BFO, PCA, and LDA. Each subfigure highlights the performance of these models under a specific feature selection approach. Fig. \ref{fig:aa} demonstrates the results with no feature selection, showing XGB achieving the highest accuracy (89.21\%), while GNB performs the lowest. Fig. \ref{fig:bb} shows the results using the BFO method, where KNN achieves the highest accuracy (90.66\%). Fig. \ref{fig:cc} uses PCA, with SVC leading in performance (86.03\%). Fig. \ref{fig:dd} incorporates LDA, where SVC achieves the best accuracy (97.93\%). These results suggest that combining MobileNetV2 with LDA and SVC yields the most accurate classification among the tested setups.

\begin{figure*}[]
    \centering
    \subfigure[]{\label{fig:ii}\includegraphics[height=3.9cm, width=4.52cm]{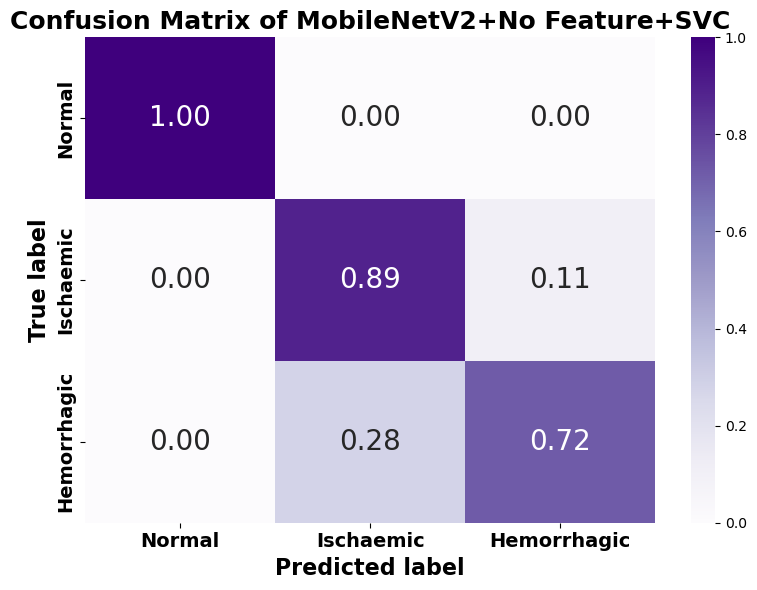}}
    \subfigure[]{\label{fig:jj}\includegraphics[height=3.9cm, width=4.52cm]{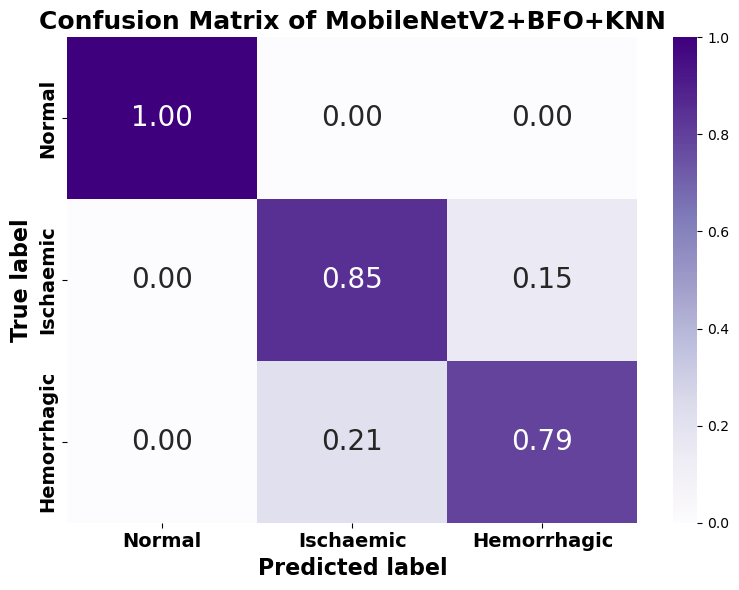}}
    \subfigure[]{\label{fig:kk}\includegraphics[height=3.9cm, width=4.52cm]{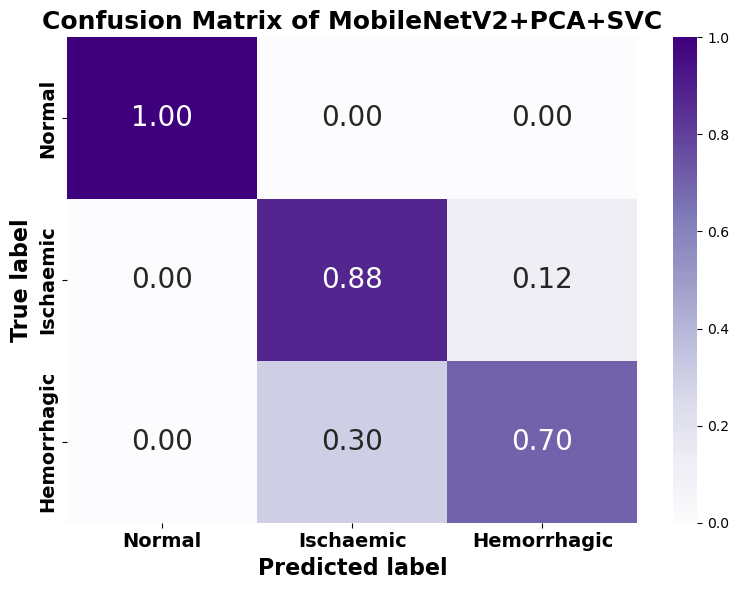}}
    \subfigure[]{\label{fig:ll}\includegraphics[height=3.9cm, width=4.52cm]{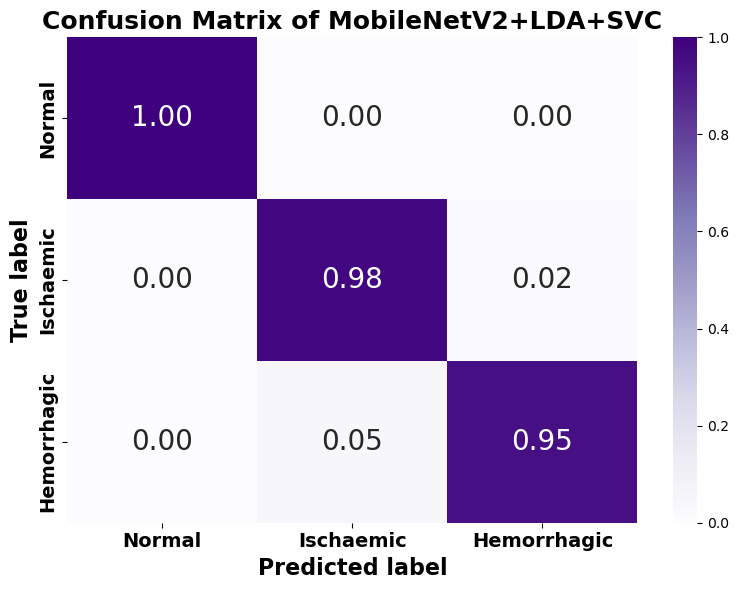}}
    \vspace{-0.4cm}
    \caption{Confusion matrices illustrating the performance of different feature selection techniques applied with MobileNetV2 and different classifiers: (a) No feature selection with Extreme Gradient Boosting, (b) Bacterial Feature Optimization with K-Nearest Neighbors, (c) Principal Component Analysis with Support Vector Classifier, and (d) Linear Discriminant Analysis with Support Vector Classifier. Each matrix shows the classification performance across three classes: Normal, Ischemic, and Hemorrhagic, with values representing the number of correctly and incorrectly classified instances.}
    \label{fig:cm}
\end{figure*}

\begin{figure*}[]
    \centering
    \subfigure[]{\label{fig:mm}\includegraphics[height=3.1cm, width=4.52cm]{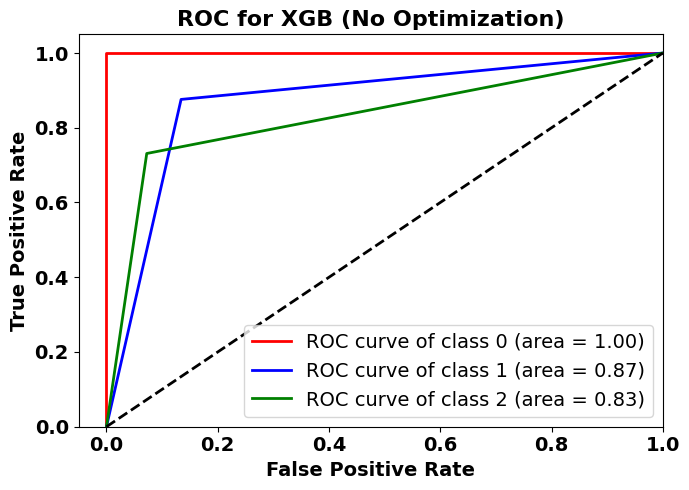}}
    \subfigure[]{\label{fig:nn}\includegraphics[height=3.1cm, width=4.52cm]{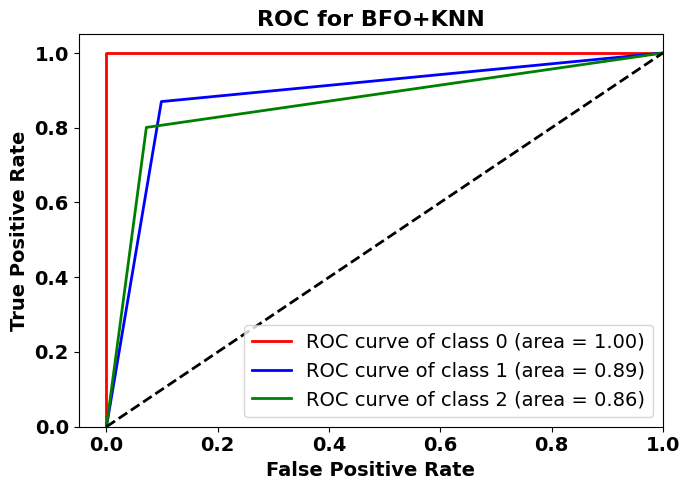}}
    \subfigure[]{\label{fig:oo}\includegraphics[height=3.1cm, width=4.52cm]{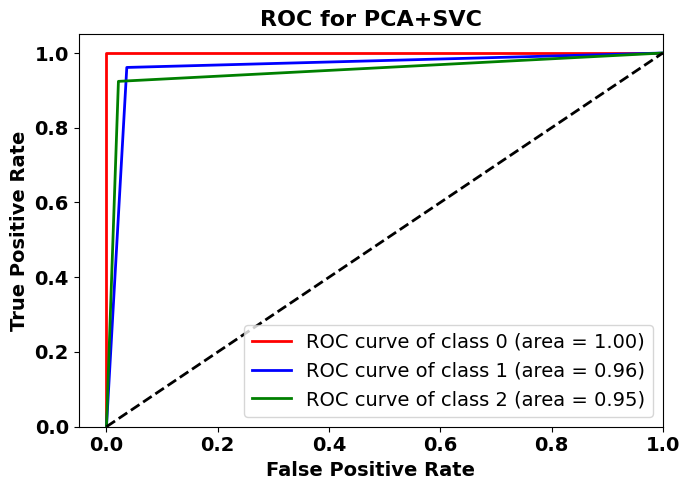}}
    \subfigure[]{\label{fig:pp}\includegraphics[height=3.1cm, width=4.52cm]{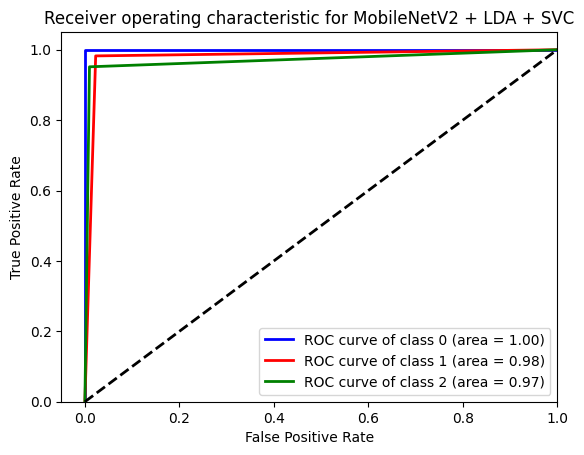}}
    \vspace{-0.4cm}
    \caption{ROC Curve illustrating the performance of different feature optimization techniques applied with MobileNetV2 and various classifiers: (a) No feature selection with SVC, (b) Bacterial Foraging Optimization with K-Nearest Neighbors, (c) Principal Component Analysis with SVC, and (d) Linear Discriminant Analysis with Support Vector Classifier. Figures show the ROC Curve across three classes: 0 (Normal), 1 (Ischemic), 2 (Hemorrhagic), with values.}
    \label{fig:rc}
\end{figure*}

Fig. \ref{fig:lc} displays learning curves that compare the performance of different feature optimization techniques applied to MobileNetV2 in terms of accuracy scores across varying training set sizes. Among the approaches, Fig. \ref{fig:hh} LDA + SVC outperforms the others, demonstrating the highest and most stable cross-validation accuracy. In contrast, Fig. \ref{fig:gg} PCA + SVC shows moderate performance but suffers from higher variance. Fig. \ref{fig:ff} BFO + KNN and Fig. \ref{fig:ee} XGB without optimization exhibit relatively lower accuracy and greater fluctuations. While all models improve with increased training data, LDA + SVC proves to be the most effective combination, offering superior and consistent classification performance.

Fig. \ref{fig:cm} illustrates confusion matrices comparing different feature optimization techniques applied to MobileNetV2 for classifying Normal, Ischemic, and Hemorrhagic cases. Among the methods, Support Vector Classifier (SVC) with Linear Discriminant Analysis (LDA) outperforms the others, achieving the highest accuracy with 98\% for the Ischemic class and 95\% for the Hemorrhagic class. Other techniques, such as Bacterial Feature Optimization (BFO) + K-Nearest Neighbors (KNN) and Principal Component Analysis (PCA) + SVC, show relatively lower performance, with misclassification rates between 11\% and 30\% in certain cases. The no-feature-selection approach with SVC demonstrates moderate classification accuracy, but LDA + SVC proves to be the best-performing combination.

Fig. \ref{fig:rc} presents ROC curves comparing different feature optimization techniques applied to MobileNetV2 with various classifiers. Among them, Fig. \ref{fig:pp}, the LDA + SVC combination, demonstrates the best overall performance, achieving the highest AUC values across all three classes. Compared to Fig. \ref{fig:oo}, the PCA + SVC combination, and Fig. \ref{fig:nn}, the BFO + KNN approach, the LDA-based method exhibits superior class separability, particularly for class 1 and class 2, which are typically more challenging to classify. On the other hand, Fig. \ref{fig:mm}, XGBoost without optimization, struggles with lower AUC scores, especially for class 2 and 3. These results indicate that LDA effectively enhances feature discrimination, making its combination with SVC the most reliable approach for our brain stroke classification task.

Table \ref{tab:tc} compares the computational complexity of various deep learning techniques for stroke detection, analyzing both time and space requirements. Among the models, InceptionV3+SVC is the fastest, with a time complexity of 132ms ± 10.2ms, while Xception+KNN demands the most memory (~1962.64 MiB PM). DenseNet201+SVC has the lowest peak memory usage (1493.54 MiB) but suffers from a higher inference time (309ms ± 128ms). MobileNetV2+SVC achieves a strong balance with 186ms ± 48.3ms time complexity and a moderate memory footprint (PM: 1546.84 MiB, INC: 569.09 MiB). Beyond computational efficiency, MobileNetV2+SVC stands out as the best model in terms of overall performance, excelling in accuracy and other classification metrics
\begin{table}[htbp]
\centering
\scriptsize
\caption{Complexity of Various Techniques for Stroke Detection}
\begin{tabular}{ccccccc}
\hline
\textbf{Technique}  &  \textbf{Time Complexity} & \textbf{Space Complexity}        \\ \hline
DenseNet201+SVC &          309ms ± 128ms            & MP: 1493.54 MiB, INC: 557.73 MiB \\
InceptionV3+SVC     &  132ms ± 10.2ms           & PM: 1789.20 MiB, INC: 705.63 MiB \\
Xception+KNN    &  177ms ± 52ms             & PM: 1962.64 MiB, INC: 792.05 MiB \\
MobileNetV2+SVC &  186ms ± 48.3ms             & PM: 1546.84 MiB, INC: 569.09 MiB \\
ResNet50+SVC        &  305ms ± 54.4ms           & PM: 1960.76 MiB, INC: 791.12 MiB \\ \hline
\end{tabular}
\label{tab:tc}
\end{table}
\begin{table*}[htbp]
\centering
\scriptsize
\setlength{\tabcolsep}{8pt}
\renewcommand{\arraystretch}{1.1}
\caption{Comparison with State-of-the-Art Methods for Brain Stroke Diagnosis Using Computed Tomography Images}
\label{tab:cmp}
\begin{tabular}{@{}lccccccc@{}} 
\toprule
\textbf{Ref.} & \textbf{Dataset} & \textbf{Class} & \textbf{Accuracy} & \textbf{Feature Optimization} & \textbf{Cross validation} & \textbf{Time\&Space Complexity} & \textbf{Computational Cost} \\ \midrule
\textit{A. Abumihsan et al.} \cite{abumihsan} & 2000 \& 2501 & 2 & 99.21\%  & \ding{51} & \ding{55} & \ding{55} & Moderate \\
\textit{M.M Hossain et al.} \cite{hossainmaruf}  & 770 \& 2501 & 2 & 96.61\%  & \ding{55}   & \ding{51}  & \ding{55}  & High  \\
\textit{I. Tahyudin et al.} \cite{tahyudin2025high} & 2501 & 2 & 95.00\%  & \ding{55} & \ding{55} & \ding{55} & High \\
\textit{UmaMaheswaran et al.} \cite{umamaheswaran2024enhanced} & 2501 &  2 & 97.00\%  & \ding{51} & \ding{55} & \ding{55} & Low \\
\textit{M.A. Saleem et al.} \cite{saleem} &    2501 & 2
 & 96.50\%  & \ding{55}  & \ding{51}  & \ding{55}  & High \\
\textit{M. Sabir et al.} \cite{sabir}  & 10000 & 2 & 96.50\%  & \ding{55}  & \ding{55}  & \ding{55}  & High \\
\textit{S. Prasher et al.} \cite{prasher2024brain} & 2501 & 2 & 98.72\%  & \ding{55}  & \ding{55}  & \ding{55}  & High \\
\textit{C.D. Kulathilake et al.} \cite{kulathilake} & 4767  & 2 &  98.02\% & \ding{55} & \ding{55} & \ding{55} & High \\
\textit{R. Raj et al.} \cite{raj2023strokevit} & 2431 & 2 &  92.00\% & \ding{51} & \ding{51} & \ding{55} & High \\
\textit{A. Diker et al.} \cite{diker}  & 2501 & 2 & 97.06\% & \ding{55} & \ding{55} & \ding{55} & High \\
\textit{\textbf{{[}This Study{]}}} & \textbf{3819} & 3 & \textbf{97.93\%} & \ding{51}  & \ding{51}  & \ding{51}  &  Low \\ \bottomrule
\end{tabular}
\end{table*}
while maintaining a good trade-off between speed and memory usage. Its relatively low computational cost makes it a practical choice for real-time stroke detection applications.

\subsection{Discussion and Limitations}
In this study, we proposed a feature extraction-based approach utilizing pre-trained deep learning models combined with various classifiers and feature optimization techniques to enhance classification performance. Instead of relying on computationally expensive end-to-end deep learning models, we leveraged transfer learning to extract meaningful features from brain imaging data, which were then fed into traditional machine learning classifiers for stroke classification. Our extensive experimentation with multiple model-classifier combinations revealed that MobileNetV2+LDA+SVC was the best-performing model. This combination achieved the highest accuracy, precision, recall, and F1-score while maintaining a strong trade-off between computational efficiency and classification performance. Additionally, the model demonstrated robustness across different feature selection techniques, further validating its reliability for stroke detection. Compared to other deep learning models, MobileNetV2 is a lightweight architecture, making it a practical choice for real-world applications where computational resources may be limited.

We compared our findings with prior research on brain stroke detection using computed tomography (CT) images; our model outperformed most existing studies in this domain. While some studies, such as \cite{abumihsan, prasher2024brain, kulathilake} used Hybrid CBAM, EfficientNet-B0, and DenseNet201, reported slightly higher accuracy. it is important to note that these models likely employed end-to-end training strategies that require extensive computational resources and large-scale datasets for optimal performance. In contrast, our feature extraction-based method offers a more efficient and scalable solution, as it eliminates the need for training deep models from scratch. Additionally, many of the high-performing models in prior studies do not report computational complexity, making it difficult to assess their feasibility for deployment in real-time clinical settings. By incorporating feature optimization techniques such as PCA, LDA, and BFO, our approach not only enhances classification accuracy but also reduces the risk of overfitting and improves generalization across different dataset sizes. Another key advantage of our model is its ability to balance high classification performance with computational efficiency, making it a viable option for deployment in medical imaging systems and resource-constrained environments.

While our proposed model achieves strong performance, there are certain limitations, since we used a feature extraction-based approach rather than fine-tuning deep learning models, the model’s adaptability to highly diverse datasets might be somewhat limited. Our study did not explicitly consider factors such as noisy, low-resolution, or incomplete medical imaging data, which are common in real-world clinical environments.

\section{Conclusion}
This study explored the use of deep feature extraction combined with machine learning classifiers and optimization techniques to develop an improved stroke detection system. Through extensive experimentation, we identified MobileNetV2+SVC+LDA as the most effective model, achieving a high classification accuracy of 97.93\%, while maintaining computational efficiency. The proposed approach effectively balances performance, interpretability, and resource constraints, making it a suitable choice for stroke classification tasks. Our findings demonstrate the advantages of using feature extraction-based models over computationally expensive end-to-end deep learning frameworks. By leveraging Linear Discriminant Analysis (LDA) for dimensionality reduction and Support Vector Classifier (SVC) for classification, our system enhances both accuracy and inference speed. The results confirm that lightweight architectures like MobileNetV2, when paired with appropriate classifiers and optimization techniques, can deliver state-of-the-art performance without excessive computational demands. Future research could explore self-supervised learning and transformer-based architectures, which have shown promise in medical imaging tasks. Additionally, incorporating multi-modal data sources, such as clinical history and radiological findings, could improve diagnostic accuracy. Expanding the dataset with more diverse and real-world stroke cases will further validate the model’s generalizability. Finally, developing a real-time deployment system with interpretability features will enhance clinical adoption, aiding medical professionals in fast and accurate stroke diagnosis.

\section*{Data Availability}
The source code and the data used in this research will be made available upon request.

\subsection*{\textbf{Acknowledgements}}
The authors would like to express their sincere gratitude to Saddam Hossain from Pakundia Upazila Health Complex for his valuable assistance and expert guidance in classifying and integrating the CT scan datasets used in this study. His insights were instrumental in constructing a reliable three-class dataset for brain stroke detection. We deeply appreciate his support and contribution to this research.

\section*{Author Contributions}
\textbf{Conceptualization:} Md. Sabbir Hossen, Eshat Ahmed Shuvo, Pabon Shaha; \textbf{Methodology:} Eshat Ahmed Shuvo, Md. Sabbir Hossen, Shibbir Ahmed Arif; \textbf{Formal analysis and investigation:} Shibbir Ahmed Arif, Md. Saiduzzaman; \textbf{Writing - original draft preparation:} Md. Sabbir Hossen; \textbf{Writing - review and editing:} Md. Sabbir Hossen, Pabon Shaha, Mostofa Kamal Nasir; \textbf{Resources:} Shibbir Ahmed Arif, Md. Saiduzzaman; \textbf{Supervision:} Mostofa Kamal Nasir, Pabon Shaha. 

\section*{Declaration of the Writing Process} 
The authors used Quilbot, Grammarly, and ChatGPT to enhance the language and readability of the work during its preparation. After utilizing these tools, the authors assumed full responsibility for the content, thoroughly reviewing and making any necessary revisions before publication.

\section*{Funding}
None reported.

\subsection*{\textbf{Declaration of Competing Interest}}
The authors declare that they have no competing interests.

\balance

\bibliographystyle{unsrt}
\bibliography{ref}

\end{document}